\documentclass[pdflatex,sn-chicago]{sn-jnl}% Chicago-based Humanities Reference Style

\usepackage{graphicx}%
\usepackage{multirow}%
\usepackage{amsmath,amssymb,amsfonts}%
\usepackage{amsthm}%
\usepackage{mathrsfs}%
\usepackage[title]{appendix}%
\usepackage{xcolor}%
\usepackage{textcomp}%
\usepackage{manyfoot}%
\usepackage{booktabs}%
\usepackage{algorithm}%
\usepackage{algorithmicx}%
\usepackage{algpseudocode}%
\usepackage{listings}%
\usepackage{subcaption}
\usepackage{longtable}

\usepackage{amsmath, amssymb, amsfonts}
\usepackage{comment}
\newcommand{\vect}[1]{\boldsymbol{#1}}
\usepackage{tabularx}

\setcitestyle{aysep={}} % Removes the comma between author and year

\theoremstyle{thmstyleone}%
\theoremstyle{thmstyletwo}%

\theoremstyle{thmstylethree}%

\begin{document}

\title[Article Title]{CNNs, RNNs and Transformers in Human Action Recognition: A Survey and a Hybrid Model}


\author*[1]{\fnm{Khaled}\sur{Alomar}}\email{kaa1u20@soton.ac.uk}

\author[1]{\fnm{Halil Ibrahim} \sur{Aysel}}\email{hia1v20@soton.ac.uk}

\author[1]{\fnm{Xiaohao} \sur{Cai}}\email{x.cai@soton.ac.uk}

\affil[1]{\orgdiv{Electronics and Computer Science}, \orgname{University of Southampton}, \orgaddress{\city{Southampton}, \postcode{SO17 1BJ}, \country{United Kingdom}}}

%\affil[2]{\orgdiv{Department}, \orgname{Organization}, \orgaddress{\street{Street}, \city{City}, \postcode{10587}, \state{State}, \country{Country}}}

%\affil[3]{\orgdiv{Department}, \orgname{Organization}, \orgaddress{\street{Street}, \city{City}, \postcode{610101}, \state{State}, \country{Country}}}


\abstract{Human action recognition (HAR) encompasses the task of monitoring human activities across various domains, including but not limited to medical, educational, entertainment, visual surveillance, video retrieval, and the identification of anomalous activities. Over the past decade, the field of HAR has witnessed substantial progress by leveraging convolutional neural networks (CNNs) and recurrent neural networks (RNNs) to effectively extract and comprehend intricate information, thereby enhancing the overall performance of HAR systems. Recently, the domain of computer vision has witnessed the emergence of Vision Transformers (ViTs) as a potent solution. The efficacy of Transformer architecture has been validated beyond the confines of image analysis, extending their applicability to diverse video-related tasks. Notably, within this landscape, the research community has shown keen interest in HAR, acknowledging its manifold utility and widespread adoption across various domains. This article aims to present an encompassing survey that focuses on CNNs and the evolution of RNNs to ViTs given their importance in the domain of HAR. By conducting a thorough examination of existing literature and exploring emerging trends, this study undertakes a critical analysis and synthesis of the accumulated knowledge in this field. Additionally, it investigates the ongoing efforts to develop hybrid approaches. Following this direction, this article presents a novel hybrid model that seeks to integrate the inherent strengths of CNNs and ViTs.}




\maketitle

\section{Introduction}\label{sec:sec1}

% What is HAR and what is its applications
% The importance of HAR
Human action recognition (HAR) focuses on the classification of the specific actions exhibited within a given video. On the other hand, action detection and segmentation focus on the precise localization or extraction of individual instances of actions from video content \citep{ulhaq2022vision}. The capacity of deep learning models to effectively capture the spatial and temporal complexities inherent in video representations plays a vital role in the recognition and understanding of actions.

% CNNs in HAR

% Vision Transformers in HAR
Over the preceding decade, a considerable amount of research has been dedicated to the thorough investigation of action recognition, resulting in an extensive collection of review articles and survey papers addressing the topic \citep{pareek2021survey, sun2022human, kong2022human}. However, it is worth noting that a predominant focus of these scholarly works has been placed on the examination and evaluation of convolutional neural networks (CNNs) and traditional machine learning models within the realm of action recognition.

The advent of Transformer architecture \citep{vaswani2017attention} has sparked a paradigm shift in deep learning.
%leading researchers to swiftly adopt and adapt it to enhance the accuracy and efficiency of various computer vision tasks including action recognition. 
%
By employing a multi-head self-attention layer, the Transformer model computes sequence representations by effectively aligning words within the sequence with other words in the same sequence \citep{ulhaq2022vision}. This approach outperforms traditional convolutional and recursive operations in terms of representation quality while utilizing fewer computational resources. As a consequence, the Transformer architecture diverges from conventional convolutional and recursive methods, favoring a more focused utilization of multiple processing nodes. The incorporation of multi-head attention allows the Transformer model to collectively learn a range of representations from diverse perspectives through the collaboration of multiple attention layers. Inspired by Transformers, many natural language processing (NLP) tasks have achieved remarkable performance, reaching human-level capabilities, as exemplified by models such as GPT \citep{brown2020language} and BERT \citep{devlin2018bert}.

The remarkable achievements of Transformers in handling sequential data, particularly in the domain of NLP, have prompted the exploration and advancement of Vision Transformer (ViT) \citep{dosovitskiy2020image} (a special Transformer for computer vision tasks). ViTs have demonstrated comparable or even superior performance compared to CNNs in the context of image recognition tasks, especially when operating on vast datasets such as ImageNet \citep{han2022survey, lin2022survey, khan2022Transformers}. This observation signifies a noteworthy shift in the field, wherein ViTs possess the potential to supplant the established dominance of CNNs in computer vision, mirroring the displacement witnessed in the case of recurrent neural networks (RNNs) \citep{ulhaq2022vision}. The achievements of Transformer models have engendered considerable scholarly interest within the computer vision research community, prompting rigorous exploration of their efficacy in pure computer vision tasks.

The natural progression in the advancement of ViTs has led to the logical exploration of video recognition tasks. Unlike image recognition, video recognition focuses on the complex challenge of identifying and understanding events within video sequences, including the recognition of human actions. 
Consequently, there is a compelling need for a recent review that comprehensively examines the state-of-the-art research including ViTs and hybrid models in addition to CNNs and RNNs for HAR. Such a review would serve as a crucial guiding resource to shape the future research directions with Transformer and CNN-Transformer hybrid architectures beside CNNs which previously were seen as unique and influential models for HAR.
The main contributions of this paper is as follows.
\begin{itemize}
\item We present a thorough review of the CNNs, RNNs and ViTs. This review examines the evolution from traditional methods to the latest advancements in neural network architectures.

\item We present an extensive examination of existing literature related to HAR.

\item We propose a novel hybrid model integrating the strengths of CNNs and ViTs. In addition, we provide a detailed performance comparison of the proposed hybrid model against existing models. The analysis highlights the model's efficacy in handling complex HAR tasks with improved accuracy and efficiency.

\item We also discuss emerging trends and the future direction of HAR technologies, emphasizing the importance of hybrid models in enhancing the interpretability and robustness of HAR systems.
\end{itemize}
These contributions enrich the understanding of the current state and future prospects of HAR, proposing innovative approaches and highlighting the importance of integrating different neural network architectures to advance the field. 

The paper is structured as follows. Section \ref{sec:sec2} delves into the background, covering foundational concepts and technologies crucial to HAR, including CNNs, RNNs and ViTs, highlighting the chronological evolution of HAR deep learning technologies. Section \ref{sec:sec3} thoroughly reviews related HAR works with a brief discussion. A novel hybrid model combining CNNs and ViTs is proposed in Section \ref{sec:sec4}, including the details  of the experimental setup and the results. Section \ref{sec:sec5} discusses the challenges and their implications for future directions in HAR. Finally, Section \ref{sec:sec6} concludes the paper.

%-------------------------
\section{Background}
\label{sec:sec2}
This section provides a chronological and technical overview of three fundamental types of neural networks: CNNs, RNNs, and Transformers. CNNs, introduced in the late 1980s, revolutionized image processing by leveraging local connectivity and shared weights to efficiently detect spatial hierarchies in data. As the field progressed, RNNs emerged in the 1990s, addressing the need for modeling sequential data through their ability to maintain temporal dependencies across sequences. The advent of Transformers in 2017 marked a paradigm shift by utilizing self-attention mechanisms to capture global relationships in data more effectively, thereby enhancing performance in a wide array of tasks beyond sequential data. This background section will delve into the technical intricacies and evolutionary trajectory of these architectures, highlighting their contributions and transitions in the realm of deep learning.

\subsection{CNNs}
%CNNs have undergone significant evolution from handling primarily spatial tasks to managing spatio-temporal tasks. Here is a detailed overview of this transition.

The evolution of CNNs has been remarkable since their introduction in the 1980s. Originally, CNNs were designed to process static images, primarily focusing on spatial recognition tasks such as object and pattern recognition. The initial idea was to build layers of convolutional filters that would apply various operations to the image to extract features like edges, textures, and shapes. This structure proved highly effective for tasks like image classification, object detection, image segmentation and more in computer vision.  

The Neocognitron \citep{fukushima1980neocognitron}, developed by Kunihiko Fukushima, presented an early example of neural networks incorporating convolutional operations for image processing, setting the foundations for subsequent progress. Shortly after, Yann LeCun and collaborators introduced LeNet-5 \citep {lecun1998gradient}, a key architecture designed for handwritten digit recognition, showcasing the effectiveness of convolutional layers in pattern recognition tasks. The progress of CNNs reached a turning point in the mid-2010s with the introduction of models like AlexNet \citep{krizhevsky2012imagenet}, showcasing their potential in image classification tasks. Alongside architectural innovations, this milestone was achieved thanks to access to large datasets, notably, ImageNet \citep{deng2009imagenet}, and computational improvements, including the rise of graphics processing units (GPUs) for parallel computing. Large-scale datasets provided the diversity and complexity necessary for training deep networks, while enhanced computational power accelerated the training of sophisticated CNN architectures.

The architectural enhancements, large datasets, and increased computational capabilities helped CNNs to be a cornerstone in deep learning methodologies, extending their applications beyond image processing to various domains. Notable architectures like VGGNet \citep{simonyan2014two}, distinguished by its uniform design and small convolutional filters, GoogLeNet \citep{szegedy2015going}, with its inception modules for capturing features at different scales efficiently, and ResNet \citep{he2016deep}, which introduced residual learning for training very deep networks, have further enriched the landscape of CNNs.

\subsubsection{Spatio-Temporal CNNs}
As CNNs excelled in spatial tasks, researchers began exploring their potential in handling temporal data, such as video and time-series analysis. The challenge was to incorporate the dimension of time into the inherently spatial architecture of CNNs.
To address this task, spatio-temporal CNNs were developed. These networks extend traditional CNN architectures by adding a temporal component to analyze dynamic behaviors across time frames. Several approaches have been utilized and main types are as follows.

3D convolution involves extending the 2D kernels to 3D, allowing the network to perform convolution across both spatial and temporal dimensions. This approach is directly applied to video data where the third dimension represents time \citep{hara2018can, tran2015learning}.
The two-stream CNNs involve running two parallel CNN streams: one for spatial processing of individual frames and another for temporal processing, usually of optical flow, which captures motion between frames \citep{simonyan2014two, feichtenhofer2016convolutional}.
RNNs with CNNs aim to combine CNNs for spatial processing with RNNs like long short-term memory (LSTM) or  gated recurrent unit (GRU) to handle temporal dependencies. This hybrid model leverages CNNs' ability to extract spatial features and RNNs' capacity to manage temporal sequences effectively \citep{yue2015beyond,donahue2015long}.

\subsection{From Vanilla RNN to Attention-Based Transformers}
This section explores the evolution from RNNs to the Transformers, highlighting the progression in handling time series and sequence data. Initially, RNNs were the go-to deep learning technique for managing temporal tasks, effectively capturing sequential dependencies. However, the development of Transformers marked a significant leap forward, driven by a series of iterative improvements and optimizations that built upon the limitations of RNNs. Transformers, with their focus on NLP, introduced a novel attention mechanism that allows for more efficient and scalable processing of sequential data. By examining the foundational RNN techniques and the subsequent enhancements leading to the Transformer architecture, this section elucidates the transformative journey from traditional RNN models to the sophisticated attention-based frameworks that now dominate the field.

We firstly establish common notations for RNN architectures including vanilla RNNs, LSTM and GRU to streamline discussions in subsequent sections. In these architectures, each iteration involves a cell that sequentially processes an input embedding $\vect{x}_{t} \in \mathbb{R}^{n_x}$ and retains information from the previous sequence through the hidden state $\vect{h}_{t-1} \in \mathbb{R}^{n_h}$ using weight matrices $\vect{W} \in \mathbb{R}^{n_h\times n_h}$ and $\vect{U} \in \mathbb{R}^{n_h\times n_x}$. The $\vect{W}$-like matrices encompass all weights related to hidden-to-hidden connections, while $\vect{U}$-like matrices encompass all weight matrices related to input-to-hidden connections. Additionally, bias terms are represented by $\vect{b}$-like vectors. Each cell produces a new hidden state $\vect{h}_{t} \in \mathbb{R}^{n_h}$ as its output.

\subsubsection{Vanilla RNNs} 
\begin{figure}[htbp]
    \centering
    \begin{tabular}{ccc}
        \begin{subfigure}[b]{0.3\textwidth}
            \centering
            \includegraphics[scale=0.5]{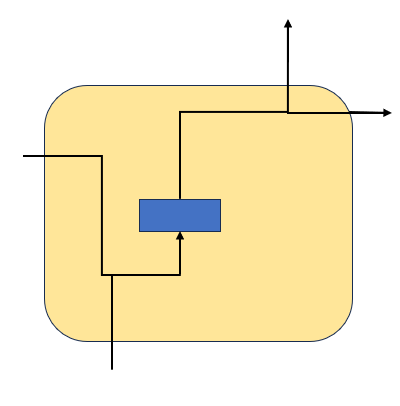}
            \put(-75,55){\scriptsize\textcolor{white}{tanh}}
            \put(-128,80){\scriptsize $\vect{h}_{t-1}$}
            \put(-86,5){\scriptsize $\vect{x}_{t}$}
            \put(-3,77){\scriptsize $\vect{h}_{t}$}
             \put(-30,110){\scriptsize $\vect{h}_{t}$}
            \caption{Vanilla RNN}
            \label{subfig:vanilla}
        \end{subfigure} 
         \hspace{0.3cm}
        &
        \begin{subfigure}[b]{0.3\textwidth}
            \centering        \includegraphics[scale=0.5]{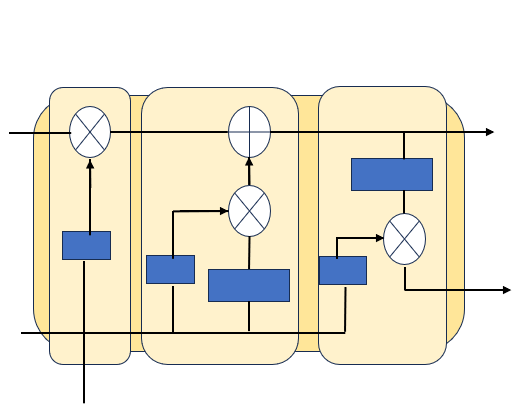}
            \put(-47,71){\scriptsize\textcolor{white}{tanh}}
            \put(-88,37){\scriptsize\textcolor{white}{tanh}}
            \put(-130,50){\scriptsize \textcolor{white}{$\sigma$}}
            \put(-105,42){\scriptsize \textcolor{white}{$\sigma$}}
            \put(-52,42){\scriptsize \textcolor{white}{$\sigma$}}
            \put(-140,105){\scriptsize Forget}
            \put(-100,105){\scriptsize Input}
            \put(-57,105){\scriptsize Output}
            \put(-170,90){\scriptsize $\vect{c}_{t-1}$}
            \put(-168,30){\scriptsize $\vect{h}_{t-1}$}
            \put(-130,0){\scriptsize $\vect{x}_{t}$}
            \put(-10,90){\scriptsize $\vect{c}_{t}$}
            \put(-1,28){\scriptsize $\vect{h}_{t}$}
            \put(-127,63){\scriptsize $\vect{f}_{t}$}
            \put(-101,55){\scriptsize $\vect{i}_{t}$}
            \put(-78,47){\scriptsize $\vect{\tilde{c}}_{t}$}
            \put(-50,57){\scriptsize $\vect{o}_{t}$}
            \caption{LSTM}
            \label{subfig:LSTM}
        \end{subfigure}
        \hspace{1cm}            
            \begin{subfigure}[b]{0.3\textwidth}
            \centering
            \includegraphics[scale=0.22]{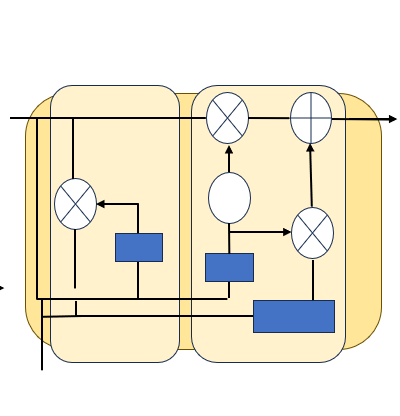}
            \put(-41,23){\scriptsize\textcolor{white}{tanh}}
            \put(-84,45){\scriptsize \textcolor{white}{$\sigma$}}
            \put(-55,40){\scriptsize \textcolor{white}{$\sigma$}}
            \put(-95,102){\scriptsize Reset}
            \put(-56,102){\scriptsize Update}
            \put(-134,93){\scriptsize $\vect{h}_{t-1}$}
            \put(0,78){\scriptsize $\vect{h}_{t}$}
            \put(-100,-2){\scriptsize $\vect{x}_{t}$}
            \put(-78,58){\scriptsize $\vect{r}_{t}$}
            \put(-64,51){\scriptsize $\vect{z}_{t}$}
            \put(-58,60){\scriptsize $1-$}
            \put(-40,38){\scriptsize $\vect{\Tilde{h}}_{t}$}\\
            \caption{GRU}
            \label{subfig:GRU}
        \end{subfigure}
         \hspace{2cm} 
    \end{tabular}
    \caption{Various types of RNN cells.}
\end{figure}
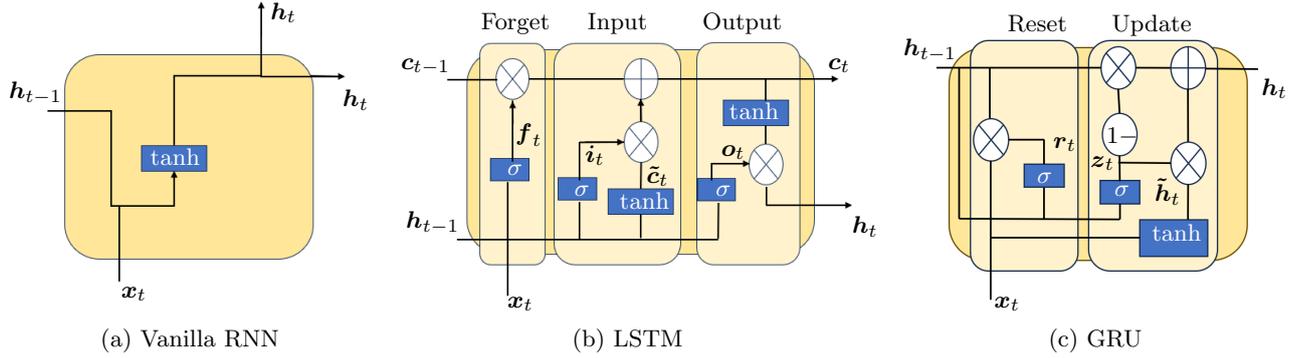

Vanilla RNNs \citep{rumelhart1985learning,  jordan1986serial} lack the presence of a cell state, relying solely on the hidden states as the primary means of memory retention within the RNN framework. The hidden state $\vect{h}_{t}$ is subsequently updated and propagated to the subsequent cell, or alternatively, depending on the specific task at hand, it can be employed to generate a prediction. Figure \ref{subfig:vanilla} illustrates the internal mechanisms of an RNN and a mathematical description of it given as
\begin{equation} \label{eq1}
\vect{h}_t=\tanh(\vect{W} \vect{h}_{t-1} + \vect{U} \vect{x}_{t} +\vect{b}),
\end{equation}
where tanh is the activation function.

Vanilla RNNs effectively incorporate short-term dependencies of temporal order and past inputs in a meaningful manner. However, they are characterized by certain limitations. Firstly, due to their intrinsic sequential nature, RNNs pose challenges in parallelized computations \citep{graves2013speech}. Consequently, this limitation can impose restrictions on the overall speed and scalability of the network. Secondly, when processing lengthy sequences, the issue of exploding or vanishing gradients may arise, thereby impeding the stable training of the network \citep{bengio1994learning}.

%----------------

\subsubsection{LSTM}    
Hochreiter and Schmidhuber (1997) introduced the LSTM cell as a solution to address the issue of long-term dependencies and to mitigate the challenge of interdependencies among successive steps \citep{hochreiter1997long}. LSTM architecture incorporates a distinct component known as the cell state $\vect{c}_{t} \in \mathbb{R}^{n_h}$, illustrated in Figure \ref{subfig:LSTM}. Analogous to a freeway, this cell state facilitates the smooth flow of information, ensuring that it can readily traverse without undergoing significant alterations.

\cite{gers2000learning} made modifications to the initial LSTM architecture by incorporating a forget gate within the cell structure. The mathematical expressions describing this modified LSTM cell are derived from its inner connections. Hence, the LSTM cell can be formally represented based on the depicted interconnections as follows.
\begin{itemize}
    \item {Forget gate} decides what information should be thrown away or kept from the cell state with the equation
    \begin{equation}
    \vect{f}_t = \sigma(\vect{W}_f \vect{h}_{t-1} + \vect{U}_{f} \vect{x}_{t} + \vect{b}_f),
          \end{equation}
where $\vect{f}_{t} \in \mathbb{R}^{n_h}$ is the output of the forget gate and $\sigma$ is the sigmoid activation function.

    \item Input gate determines which new information is added to the cell state with two activation functions defined as   
   \begin{equation}
       \vect{i}_t = \sigma(\vect{W}_i  \vect{h}_{t-1} + \vect{U}_i \vect{x}_t + \vect{b}_i),
   \end{equation}
where $\vect{i}_t \in \mathbb{R}^{n_h}$ is the output of the sigmoid activation function;  and  
\begin{equation}
            \vect{\tilde{c}}_t = \tanh(\vect{W}_{\tilde{c}} \vect{h}_{t-1} + \vect{U}_{\tilde{c}} \vect{x}_t + \vect{b}_{\tilde{c}}),
        \end{equation}
where $\vect{\Tilde{c}}_t \in \mathbb{R}^{n_h}$ is known as candidate value. After obtaining $\vect{i}_t$ and $\vect{\Tilde{c}}_t$, we can update the cell state with
    \begin{equation}
            \vect{c}_t = \vect{f}_t \odot \vect{c}_{t-1} + \vect{i}_t \odot \vect{\tilde{c}}_t,
        \end{equation}
where $\vect{c}_{t-1} \in \mathbb{R}^{n_h}$ is the previous cell state and $\odot$ is the Hadamard operator.
    
    \item Output gate determines the next hidden state based on the cell state and output gate's activity
    \begin{equation}
            \vect{o_t} = \sigma(\vect{W}_o  \vect{h}_{t-1} + \vect{U}_o \vect{x}_t + \vect{b}_o),
        \end{equation}
    where $\vect{o}_t \in \mathbb{R}^{n_h}$ is the output of the output gate. Finally the updated hidden state,
        \begin{equation}
            \vect{h}_t = \tanh( \vect{c}_t) \odot  \vect{o}_{t}
        \end{equation}      
is fed to the next iteration.
\end{itemize}

%%%%%%%%%%%%%%%%%%%%%%%%%%

To enable selective information retention, LSTM employs three distinct gates. The first gate, known as the forget gate, examines the previous hidden state $\vect{h}_{t-1}$ and the current input $\vect{x}_{t}$. It generates a vector $\vect{f}_{t}$ containing values between $0$ and $1$, determining the portion of information to discard from the previous cell state $\vect{c}_{t-1}$. The second gate, referred to as the input gate, follows a similar process to the forget gate. However, instead of discarding information, it utilizes the output $\vect{i}_{t}$ to determine the new information to be stored in the cell state based on a candidate cell state $\vect{\tilde{c}}_t$.  Lastly, the output gate employs the output $\vect{o}_{t}$ to filter the updated cell state $\vect{c}_{t}$, thereby transforming it into the new hidden state $\vect{h}_{t}$. The LSTM cell exhibits superior performance in retaining both long-term and short-term memory compared to the vanilla RNN cell. However, this advantage comes at the expense of increased complexity.

\subsubsection{GRU} The LSTM cell surpasses the learning capability of the conventional recurrent cell, yet the additional number of parameters escalates the computational load. Consequently, to address this concern, \cite{chung2014empirical} introduced the GRU, see Figure \ref{subfig:GRU}. GRU demonstrates comparable performance to LSTM while offering a more computationally efficient design with fewer weights. This is achieved by merging the cell state and the hidden state into ``reset state" resulting in a simplified architecture. Furthermore, GRU combines the forget and input gates into an ``update gate", contributing to a more streamlined computational process. For further elaboration, GRU cell incorporates two essential gates. The first gate is the reset gate, which examines the previous hidden state $\vect{h}_{t-1}$ and the current input $\vect{x}_{t}$. It generates a vector $\vect{r}_{t}$ containing values between $0$ and $1$, determining the extent to which past information in $\vect{h}_{t-1}$ should be disregarded. The second gate is the update gate, which governs the selection of information to either retain or discard when updating the new hidden state $\vect{h}_{t}$, based on the value of $\vect{r}_{t}$.

Based on the depicted information in Figure \ref{subfig:GRU}, the mathematical expressions governing the behavior of the GRU cell can be expressed as follows.
\begin{itemize}

    \item Update gate decides how much of the past information needs to be passed along with
    \begin{equation}
        \vect{z}_t = \sigma(\vect{W}_z \vect{h}_{t-1} + \vect{U}_z \vect{x}_{t} + \vect{b}_z),
    \end{equation}
where $\vect{z}_t \in \mathbb{R}^{n_h}$ is the output of the update gate. The output of the reset gate $\vect{r}_t \in \mathbb{R}^{n_h}$ is obtained by
    \begin{equation}
        \vect{r}_t = \sigma(\vect{W}_r \vect{h}_{t-1} + \vect{U}_r \vect{x}_{t} + \vect{b}_r).
    \end{equation}
A candidate activation for the subsequent step is
    \begin{equation}
        \vect{\tilde{h}}_t = \tanh(\vect{W}_{\tilde{h}} (\vect{r}_t \odot \vect{h}_{t-1}) + \vect{U}_{\tilde{h}} \vect{x}_t + \vect{b}_{\tilde{h}})
    \end{equation}
where $\vect{\tilde{h}}_t  \in \mathbb{R}^{n_h}$.

    \item The final activation is a blend of the previous hidden state and the candidate activation, weighted by the update gate, i.e.,
        \begin{equation}
        \vect{h}_t = \vect{z}_t \odot \vect{\tilde{h}}_t + (1 - \vect{z}_t) \odot \vect{h}_{t-1}
    \end{equation}
where $\vect{h}_t \in \mathbb{R}^{n_h}$ is the updated hidden state. This mechanism allows the GRU to effectively retain or replace old information with new information.

\end{itemize}

\begin{figure}[H]
\centering
\includegraphics[scale=0.5]{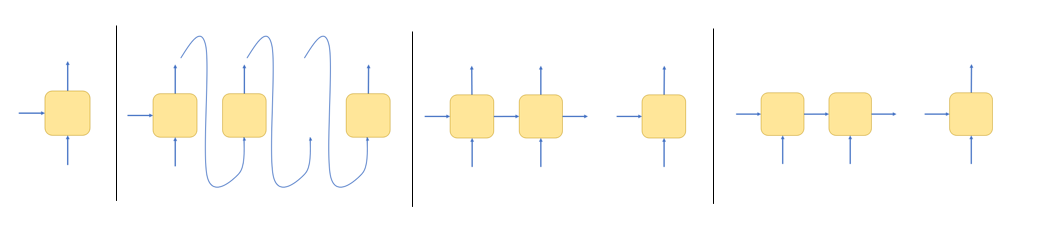}
 \put(-395,8){\scriptsize (a) One-to-one}
 \put(-395,49){\scriptsize $h_{0}$}
 \put(-370,75){\scriptsize $y_{1}$}
 \put(-370,25){\scriptsize $x_{1}$}
 \put(-320,8){\scriptsize (b) One-to-many}
 \put(-345,53){\scriptsize $h_{0}$}
 \put(-328,25){\scriptsize $x_{1}$}
 \put(-331,72){\scriptsize $y_{1}$}
 \put(-307,72){\scriptsize $y_{2}$}
 \put(-255,72){\scriptsize $y_{t}$}
 \put(-277,45){\scriptsize ...}
 \put(-210,8){\scriptsize (c) Many-to-many}
 \put(-233,53){\scriptsize $h_{0}$}
 \put(-217,25){\scriptsize $x_{1}$}
 \put(-191,25){\scriptsize $x_{2}$}
 \put(-143,25){\scriptsize $x_{t}$}
 \put(-217,72){\scriptsize $y_{1}$}
 \put(-192,72){\scriptsize $y_{2}$}
 \put(-145,72){\scriptsize $y_{t}$}
 \put(-168,49){\scriptsize ...}
 \put(-100,8){\scriptsize (d) Many-to-one}
 \put(-116,53){\scriptsize $h_{0}$}
 \put(-29,72){\scriptsize $y_{t}$}
 \put(-100,25){\scriptsize $x_{1}$}
 \put(-75,25){\scriptsize $x_{2}$}
 \put(-29,25){\scriptsize $x_{t}$}
 \put(-52,50){\scriptsize ...}

\caption{Types of RNN structures based on input-output pairs.}

\label{fig:io_types}
\end{figure}
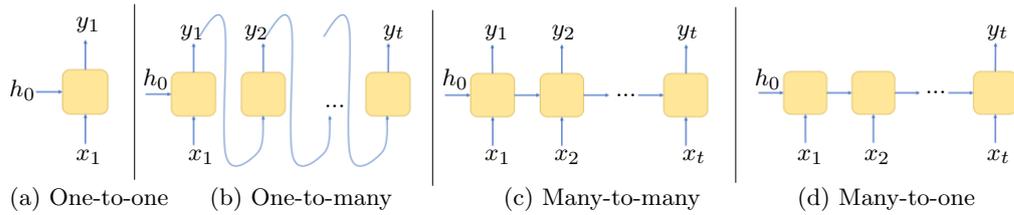

\subsubsection{Types of RNNs} RNNs were created with an internal memory mechanism that allows them to store and use information from previous outputs. This unique trait enables RNNs to retain important contextual information over time, enabling reasoned decision-making based on past results. There are four types of popular RNN variants that each serve different purposes across a variety of applications, see Figure \ref{fig:io_types}.

\begin{comment}
\begin{figure}[htbp]
    \centering
    \begin{tabular}{cccc}
        \begin{subfigure}[b]{0.2\textwidth}
            \centering
            \includegraphics[scale=0.25]{images/One-to-one.png}
            \caption{one-to-one}
            \label{subfig:one-to-one}
        \end{subfigure} 
        &
        \begin{subfigure}[b]{0.2\textwidth}
            \centering        \includegraphics[scale=0.25]{images/One-to-many.png}
            \put(-40,59){\scriptsize\textcolor{white}{Tanh}}
           \label{subfig:one-to-many}
            \caption{one-to-many}
            \label{subfig:one-to-many}
        \end{subfigure}
        \hspace{1cm}
                \begin{subfigure}[b]{0.25\textwidth}
            \centering
            \includegraphics[scale=0.25]{images/Many-to-one.png}
            \caption{many-to-one}
            \label{subfig:many-to-one}
        \end{subfigure}
         \hspace{0.5cm}
        \begin{subfigure}[b]{0.25\textwidth}
            \centering
            \includegraphics[scale=0.25]{images/Many-to-many E.png}
            \caption{many-to-many}
            \label{subfig:many-to-many}
        \end{subfigure}
    \end{tabular}
    \caption{Overall caption for the table}
\end{figure}
\end{comment}

The one-to-one is considered the simplest form of RNNs, where a single input corresponds to a single output. It operates with fixed input and output sizes, functioning similarly to a standard neural network. 
One-to-many represents a specific category of RNNs that is characterized by its ability to produce multiple outputs based on a single input provided to the model. This type of RNN is particularly useful in applications like image captioning, where a fixed input size results in a series of data outputs.
Many-to-one RNNs merge a sequence of inputs into a single output through a series of hidden layers that learn relevant features. An illustrative instance of this RNN type is sentiment analysis, where the model analyzes a sequence of text inputs and produces a single output indicating the sentiment expressed in the text.

\begin{figure}[htbp]
    \centering
    \begin{tabular}{ccc}
        \begin{subfigure}[b]{0.35\textwidth}
            \centering
            \includegraphics[scale=0.45]{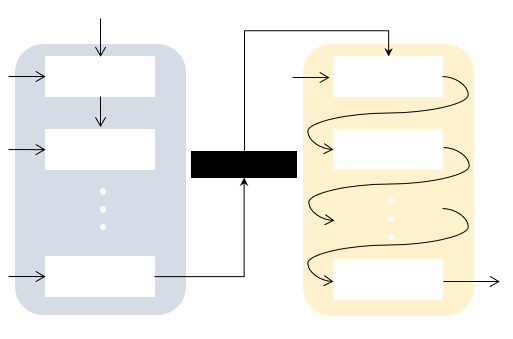}
             \put(-145,111){\scriptsize $\vect{h}_{0}$}
            \put(-183,87){\scriptsize $\vect{x}_{0}$}
            \put(-150,86){\scriptsize RNN}
            \put(-183,63){\scriptsize $\vect{x}_{1}$}
            \put(-150,62){\scriptsize RNN}
            \put(-183,19){\scriptsize $\vect{x}_{t}$}
            \put(-150,18){\scriptsize RNN}
            \put(-95,57){\scriptsize \color{white}$\vect{h}_{t}$}
             \put(-88,86){\scriptsize sos}
             \put(-52,86){\scriptsize RNN}
             \put(-45,73){$\downarrow$}
             \put(-52,61){\scriptsize RNN}
            \put(-2,18){\scriptsize eos}
             \put(-52,17){\scriptsize RNN}
            \caption{Without attention}
            \label{subfig:no_attention}
        \end{subfigure} 
        \hspace{0.8cm}
        &
        \begin{subfigure}[b]{0.35\textwidth}
            \centering        \includegraphics[scale=0.45]{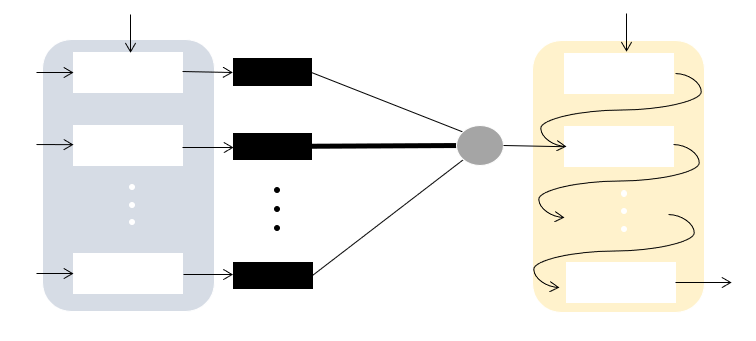}
            \put(-255,89){\scriptsize $\vect{x}_{0}$}
            \put(-223,87){\scriptsize RNN}
            \put(-215,76){$\downarrow$}
            \put(-255,64){\scriptsize $\vect{x}_{1}$}
            \put(-223,64){\scriptsize RNN}
            \put(-255,22){\scriptsize $\vect{x}_{t}$}
            \put(-223,20){\scriptsize RNN}
            \put(-50,114){\scriptsize $\vect{h}_{t}$}
            \put(-217,112){\scriptsize $\vect{h}_{0}$}
            \put(-170,88){\scriptsize\color{white}$\vect{h}_{0}$}
            \put(-132,88){\scriptsize $\alpha_{0}$}
            \put(-170,63){\scriptsize \color{white}$\vect{h}_{1}$}
            \put(-132,70){\scriptsize $\alpha_{1}$}
            \put(-170,20){\scriptsize \color{white}$\vect{h}_{t}$}
            \put(-132,45){\scriptsize $\alpha_{t}$}
            \put(-76,88){$\rightarrow$}
            \put(-90,88){\scriptsize sos}
            \put(-5,18){\scriptsize eos}
             \put(-56,89){\scriptsize RNN}
             \put(-47,75){$\downarrow$}
             \put(-56,62){\scriptsize RNN}
             \put(-56,17){\scriptsize RNN}
             \put(-97,65){\tiny \color{white} $\sum$}
            \caption{With attention}
            \label{subfig:attention}
        \end{subfigure}
        \hspace{2cm}
    \end{tabular}
    \caption{Sequence-to-sequence RNN with and without the attention mechanism. }
\end{figure}
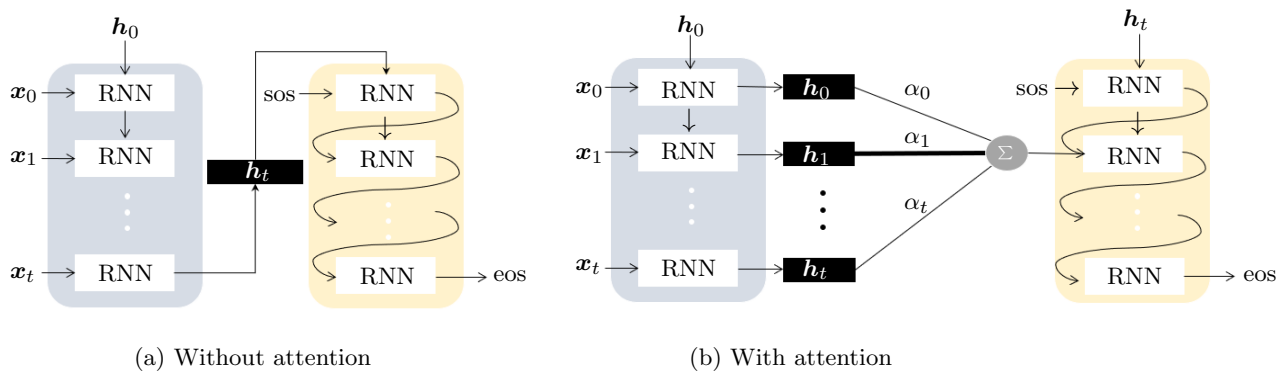

Many-to-many RNNs are employed to generate a sequence of output data from a sequence of input units. It can be categorized into two subcategories: equal size and unequal size. In the equal size subcategory, the input and output layers have the same size, see many-to-many architecture in Figure \ref{fig:io_types}c. Several research efforts have emerged to tackle the limitation of the fixed-size input-output sequences in machine translation tasks, as they fail to adequately represent real-world requirements. The unequal size subcategory can handle different sizes of inputs and outputs. A practical application of the unequal size subcategory can be observed in machine translation. In this scenario, the model generates a sequence of translated text outputs based on a sequence of input sentences. Unequal size subcategory employs an encoder-decoder architecture, where the encoder adopts the many-to-one architecture, and the decoder adopts the one-to-many architecture. One notable contribution in this area was made by \cite{kalchbrenner2013recurrent}, who pioneered the approach of mapping the entire input sentence to a vector. This work is related to the study conducted by \cite{cho2014learning}, although the latter was specifically utilized to refine hypotheses generated by a phrase-based system \citep{sutskever2014sequence}. In this architecture, the encoder component plays a crucial role in transforming the inputs into a singular vector, commonly referred to as the context. This context vector, typically with a length of 256, 512 or 1024, encapsulates all the pertinent information detected by the encoder from the input sentence, which serves as the translation target, see Figure \ref{subfig:no_attention}. Subsequently, this vector is passed on to the decoder, which generates the corresponding output sequence. It is important to note that both the encoder and decoder components in this architecture are RNNs. Different from  Figure \ref{subfig:no_attention},  Figure \ref{subfig:attention} gives the encoder-decoder architecture with attention which will be introduced in the next section.

\begin{comment}

\begin{figure}[H]
\centering
\includegraphics[width=0.9\textwidth, height=0.2\textwidth]{images/RNN_different_sizes.PNG}
\put(-420,44){\scriptsize $h_{0}$}
\put(-382,5){\scriptsize $x_{1}$}
\put(-338,5){\scriptsize $x_{2}$}
\put(-255,5){\scriptsize $x_{t}$}
\put(-220,53){\scriptsize $Context$}
\put(-159,88){\scriptsize $y_{1}$}
\put(-115,88){\scriptsize $y_{2}$}
\put(-32,3){\scriptsize $y_{t}$}
\put(-43,88){\scriptsize $<eos>$}
\put(-170,3){\scriptsize $<bos>$}
\put(-115,3){\scriptsize $y_{1}$}
\put(-297,46){\scriptsize ...}
\put(-74,46){\scriptsize ...}

\caption{Diagram of Many-to-many architecture of unequal-sized input-output (Classic SeqToSeq architecture without attention).}

\label{fig:ManyToManyUE}
\end{figure}
\end{comment}

\subsubsection{Attention}

The evolution of attention mechanisms in neural networks represents a significant advancement in the field of deep learning, particularly in tasks related to NLP and machine translation. Initially introduced by \cite{graves2013generating}, the concept of attention mechanisms was designed to enhance the model's ability to focus on specific parts of the input sequence when generating an output, mimicking the human ability to concentrate on particular aspects of a task. This foundational work laid the groundwork for subsequent developments in attention mechanisms, providing a mechanism for models to dynamically assign importance to different parts of the input data.

Building on Graves' initial concept, \cite{bahdanau2014neural} introduced the additive attention mechanism, which was specifically designed to improve machine translation. This approach computes the attention weights through a feed-forward neural network, allowing the model to consider the entire input sequence and determine the relevance of each part when translating a segment. This additive form of attention significantly improved the performance of sequence-to-sequence models by enabling a more nuanced understanding and alignment between the input and output sequences \citep{sutskever2014sequence}. Following this, \cite{luong2015effective} proposed the multiplicative attention mechanism, also known as dot-product attention, which simplifies the computation of attention weights by calculating the dot product between the query and all keys. This method not only streamlined the attention mechanism but also offered improvements in computational efficiency and performance in various NLP tasks, marking a pivotal moment in the evolution of attention mechanisms from their inception to more sophisticated and efficient variants.

%%%%%%%%%%%%%%%%%%%%%%%%%%%%%%%%%

\begin{comment}
To provide a visual and mathematical representation of the Luong's attention mechanism incorporated into the SeqtoSeq RNN architecture \citep{luong2015effective}, refer to the Figure \ref{fig:SeqtoSeqwithAttention} and the Eq \ref{eq12}.

In the Luong's approach \citep{luong2015effective}, the target hidden state at the top LSTM layers in both the encoder and decoder ($\vect{h_{t}}$) with all source hidden states of the encoder (${\vect{h_{s}}}$) are calculated to obtain the scores, as demonstrated in Eq \ref{eq12}. The attention mechanism applies a softmax function to rescale the scores, transforming them into a range of values between 0 and 1. Consequently, scores with a higher value will approach 1, while scores with a lower value will move closer to 0. By multiplying the hidden states of the encoder with the scores ($\vect{\alpha_{ts}}$) and summing the results, the context vector $\vect{c_t}$ can be obtained, see Eqs \ref{eq13} and \ref{eq14}.

\end{comment}

The central idea of the attention mechanism is to shift focus from the task of learning a single vector representation for each sentence. Instead, it adopts a strategy of selectively attending to particular input vectors in the input sequence, guided by assigned attention weights. This strategy enables the model to dynamically allocate its attention resources to the most pertinent segments of the sequence, thereby improving its capacity to process and comprehend the information more efficiently \citep{brauwers2021general}.

One possible explanation for the improvement is that the attention layer created memories associated with the context pattern rather than memories associated with the input itself, relieving pressure on the RNN model structure's weights and causing the model memory to be devoted to remembering the input rather than the context pattern \citep{hu2018novel}.

\begin{figure}[htbp]
    \centering
    \begin{tabular}{ccc}
        \begin{subfigure}[b]{0.3\textwidth}
            \centering
            \includegraphics[scale=0.5]{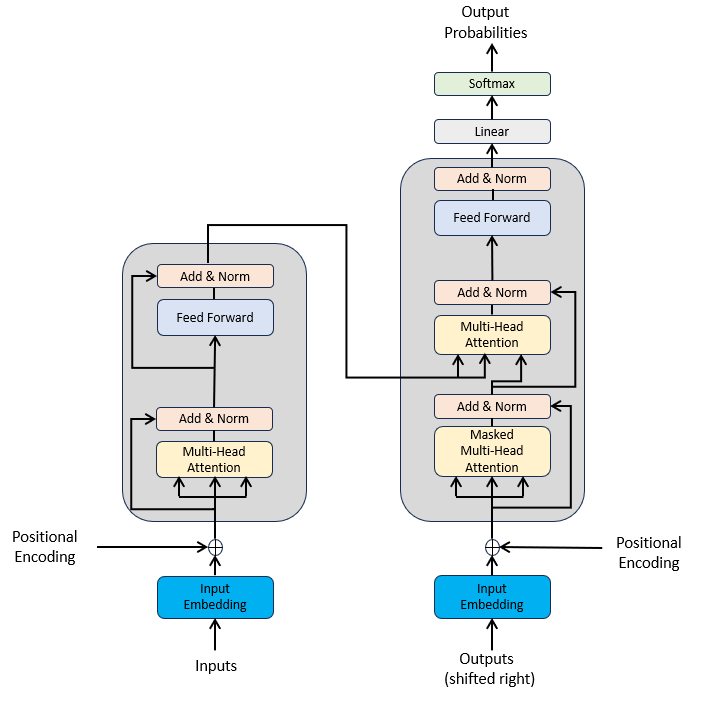}
            \caption{Transformer}
            \label{subfig:self-attention}
            
        \end{subfigure} 
         \hspace{3cm}
        &
        \begin{subfigure}[b]{0.5\textwidth}
            \centering        \includegraphics[scale=0.4]{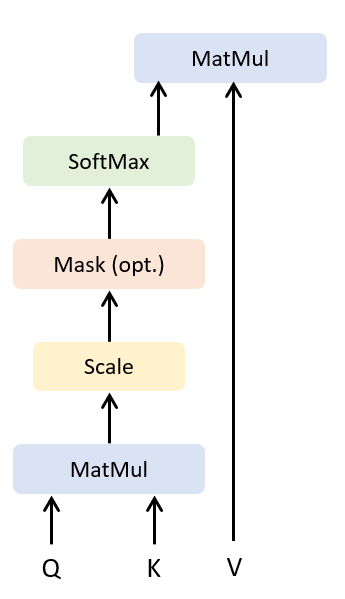}
            \caption{Self-Attention}
            \label{subfig:Transformer}
        \end{subfigure}
        \hspace{2cm}
    \end{tabular}
    \caption{Transformer architecture and its self-attention mechanism (adapted from \cite{vaswani2017attention}).}
    \label{fig:self-att}
\end{figure}

\subsubsection{Self-Attention}

To this point, attention mechanisms in sequence-transformation models have primarily relied on complex RNNs, featuring an encoder and a decoder, the most successful models in language translation yet. However, \cite{vaswani2017attention} introduced a simple network architecture known as the Transformer, see Figure \ref{fig:self-att}, which exclusively utilized attention mechanism, eliminating the need for RNNs. They introduced a novel attention mechanism called self-attention, which is also known as KQV-attention (Key, Query, and Value). This attention mechanism subsequently gained prominence as a central component within the Transformer architecture. The attention mechanism stands out due to its ability to provide Transformers with an extensive long-term memory. In the Transformer model, it becomes possible to focus on all previously generated tokens.

The embedding layer in a Transformer model is the initial stage where input tokens are transformed into dense vectors, capturing semantic information about each token's meaning and context within the text. These embeddings serve as the foundation for subsequent layers to process and understand the relationships between words in the input sequence \citep{dar2022analyzing}.

Self-attention is a mechanism that allows an input sequence to process itself in a way that each position in the sequence can attend to all positions within the same sequence. This mechanism is a cornerstone of the Transformer architecture, which has revolutionized NLP and beyond by enabling models to efficiently handle sequences of data with complex dependencies. The purpose of self-attention is to compute a representation of each element in a sequence by considering the entire sequence, thereby capturing the contextual relationships between elements regardless of their positional distance from each other. This ability to capture both local and global dependencies makes self-attention particularly powerful for tasks such as machine translation, text summarization, and sequence prediction, where understanding the context and the relationship between words or elements in a sequence is crucial \citep{vaswani2017attention}.

The mathematical formulation of self-attention involves several key steps. First, a set of query vectors $\vect{Q}= \vect{X} \vect{W}^Q$, a set of key vectors $\vect{K} = \vect{X} \vect{W}^K$, and a set of value vectors $\vect{V}= \vect{X} \vect{W}^V$  are calculated through linear transformations of the input sequence, where $\vect{X}$ is the input matrix representing embeddings of tokens in a sequence, and $\vect{W}^Q$,$\vect{W}^K$, and $\vect{W}^V$ are weight matrices for queries, keys, and values, respectively. The attention scores are then calculated by taking the dot product of the query vector with all key vectors, followed by scaling the result by the inverse square root of the dimension of the keys (say $\sqrt{d_k}$) to avoid overly large values. These scores are then passed through a softmax function to obtain the attention weights, which represent the importance of each element's contribution to the output. Finally, the output say $\vect{A}$ is computed as a weighted sum of the value vectors, i.e.,
\begin{equation}
\vect{A(\vect{Q, K, V})} = \text{softmax}(\frac{\vect{QK^\top}}{\sqrt{d_k}}) \vect{V}.
\end{equation}
This process allows the model to dynamically focus on different parts of the input sequence, enabling the extraction of rich contextual information from the sequence.

\subsubsection{Multi-Head-Attention}
Multi-head attention is an extension of the self-attention mechanism designed to allow the model to jointly attend the information from different representation subspaces at different positions \citep{vaswani2017attention}. Instead of performing a single attention function, it runs the attention mechanism multiple times in parallel. The outputs of these independent attention computations are then concatenated and linearly transformed into the expected dimension. The mathematical formulation of the multi-head attention can be described in the following steps. 
First, for the $i$-th self-attention head, find
\begin{equation}\label{eq19}
    \vect{Q}_i = \vect{X} \vect{W}_i^Q, \quad \vect{K}_i = \vect{X} \vect{W}_i^K, \quad \vect{V}_i = \vect{X} \vect{W}_i^V,
    \end{equation}
and then compute
\begin{equation}
    \vect{A}_i(\vect{Q}_i, \vect{K}_i, \vect{V}_i) = \text{softmax}\left(\frac{\vect{Q}_i \vect{K}_i^\top}{\sqrt{d_k}}\right) \vect{V}_i .
    \end{equation}
The multi-head attention is obtained by concatenating all $\vect{A}_i(\vect{Q}_i, \vect{K}_i, \vect{V}_i)$.

%Then, concatenating outputs from all heads followed by a final linear transformation,
%\begin{equation}
%    \vect{A}(\vect{Q}, \vect{K}, \vect{V}) = \text{Concat}(\vect{A}_1, \vect{A}_2, \ldots, \vect{A}_h) \vect{W}^O
%    \end{equation}
%where $\vect{W}^O$ is the weight matrix for the final linear transformation.

The multi-head attention mechanism enables the model to capture different types of information from different positions of the input sequence. By processing the sequence through multiple attention ``heads", the model can focus on different aspects of the sequence, such as syntactic and semantic features, simultaneously. This capability enhances the model's ability to understand and represent complex data, making multi-head attention a powerful component of Transformer-based architectures \citep{devlin2019bert}.

\subsection{From Transformer to Vision Transformer}

The journey from the inception of the Transformer model to the development of the ViT marks a pivotal advancement in deep learning, showcasing the adaptability of models initially designed for sequence data processing to the realm of image analysis. This transition underscores a significant shift in approach, from conventional image processing techniques to more sophisticated sequence-based methodologies.

Introduced by \cite{vaswani2017attention} through the seminal paper ``Attention Is All You Need",  the Transformer model revolutionized NLP by leveraging self-attention mechanisms. This innovation allowed for the processing of sequences of data without the reliance on recurrent layers, facilitating unprecedented parallelization and significantly reducing training times for large datasets. The Transformer's success in NLP sparked curiosity about its potential applicability across different types of data, including images, setting the stage for a transformative adaptation.

\begin{figure}[H]
\hspace{2cm}
\includegraphics[scale=0.4]{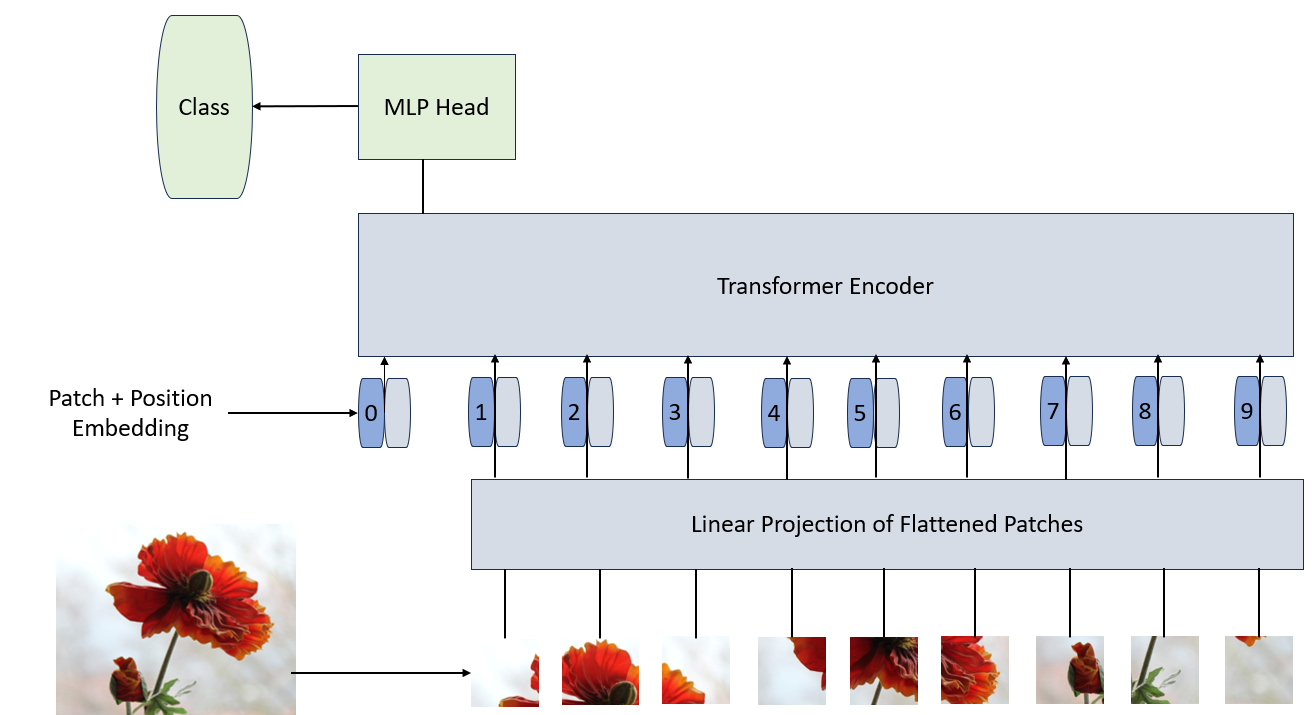}
\caption{The ViT architecture (adapted from \cite{dosovitskiy2020image}).}
\label{fig:VisionTransformer}
\end{figure}

The adaptation of Transformers for image data pivoted on a novel concept: treating images not as traditional 2D arrays of pixels but as sequences of smaller and discrete image patches. This approach, however, faced computational challenges due to the self-attention mechanism's quadratic complexity with respect to input length. The breakthrough came with the introduction of the ViT by \cite{dosovitskiy2020image}, which applied the Transformer architecture directly to images, see Figure \ref{fig:VisionTransformer}. By dividing an image into fixed-size patches and processing these patches as if they were tokens in a text sequence, ViT was able to capture complex relationships between different parts of an image using the Transformer's encoder.

The operational mechanics of ViT begin with the division of an input image into fixed-size patches, each of which is flattened and linearly transformed into a vector, effectively converting the 2D image into a 1D sequence of embeddings. To account for the lack of inherent positional awareness within the Transformer architecture, positional embeddings are added to these patch embeddings, ensuring the model retains spatial information. The sequence of embeddings is then processed through the Transformer encoder, which consists of layers of multi-head self-attention and feed-forward neural networks, allowing the model to dynamically weigh the importance of each patch relative to others for a given task.

For tasks like image classification, the output from the Transformer encoder is passed through a classification head, often utilizing a learnable ``class token" appended to the sequence of patch embeddings for this purpose. The model is trained on large datasets using backpropagation and, during inference, processes images through these steps to predict their classes.

The ViT not only demonstrates exceptional performance on image classification tasks, often surpassing CNNs when trained on extensive datasets, but also highlights the Transformer architecture's capacity to capture the global context within images. Despite its advantages, ViT's reliance on substantial computational resources for training and its need for large datasets to achieve optimal performance present challenges. Nonetheless, the development of ViT signifies a significant milestone in the application of sequence processing models to the field of computer vision, opening new avenues for research and practical applications.

%---------

%---------
The original ViT, designed for static image processing, divides images into patches and interprets these as sequences, leveraging the Transformer's self-attention mechanism to understand complex spatial relationships. Extending this model to action recognition involves adapting it to analyze video frames sequentially to capture both spatial and temporal relationships.
Several works attempted to adapt ViT in action recognition task using different methods as below.

\textit{Temporal dimension integration.} 
The integration of the temporal dimension is a fundamental step in adapting ViT for action recognition. Traditional ViT models process images as a series of patches, treating them essentially as sequences for the self-attention mechanism to analyze spatial relationships. By extending this concept to include the temporal dimension, the models can now treat videos as sequences of frame patches over time. This allows the models to capture the evolution of actions across frames. The work by \cite{bertasius2021space} highlights the potential of incorporating temporal information into Transformers, marking a significant advancement in video analysis capabilities.

\textit{Spatio-temporal embeddings.}
To effectively capture the dynamics of actions within videos, adapted ViT models generate spatio-temporal embeddings. This involves extending the traditional positional embeddings used in ViTs to also include temporal positions, thereby creating embeddings that account for both spatial and temporal information within video sequences. The discussion by \cite{arnab2021vivit} on the creation of these spatio-temporal embeddings showcases the method's effectiveness in enhancing the model's understanding of action dynamics across both space and time.

\textit{Multi-head self-attention across time.} 
The extension of self-attention mechanisms to analyze relationships between patches not just within individual frames but also across different frames is crucial for recognizing actions over time. This approach enables the model to identify relevant features and changes across the video sequences, facilitating a deeper understanding of motion and the progression of actions. The exploration by \cite{bertasius2021space} of this concept demonstrates how Transformers can be effectively adapted to capture the temporal dynamics of actions, a key aspect of video analysis.

%-----------------------------
\section{Literature Review}
\label{sec:sec3}
%-----------------------------
This section briefly recalls the most commonly used deep learning-based HAR approaches.

\subsection{CNN-Based Approaches in HAR}
This section recalls the most prominent CNN-based approaches in HAR based on the model type (i.e., the two-stream CNN, 3D CNN, and RNNs with CNNs), organized chronologically.

Deep learning was still in its early stages in 2012, and CNNs or RNNs had not yet gained significant popularity in the field of HAR. The focus was primarily on traditional machine learning approaches, such as support vector machines \citep{cortes1995support}, and handcrafted features, such as histogram of oriented gradients  \citep{dalal2005histograms} and histogram of optical flow \citep{barron1994performance}. A few studies did, nevertheless, start looking into neural networks for action recognition.

In 2014, the use of CNNs in action recognition was at a pivotal stage, marking a shift from hand-crafted feature-based methods to deep learning approaches. The key points of the use of CNNs in action recognition at that period of time are the following. (I) \textit{Emergence of deep learning:} deep learning, particularly CNNs, had started to dominate image classification tasks, thanks to their ability to learn feature representations directly from raw pixel data. This success in static images paved the way for applying CNNs to video data for action recognition. (II) \textit{Challenges in video data:} unlike 2D images, videos incorporate a third dimension which represents the temporal patterns, making action recognition more complex. CNNs had to be adapted to not only recognize spatial patterns but also capture motion information over time dimension. (III) \textit{Datasets and benchmarks:} the adoption of large-scale video datasets like UCF-101 \citep{soomro2012ucf101} and HMDB-51 \citep{kuehne2011hmdb} became more common. These datasets provided diverse sets of actions and were large enough to train deep networks. The performance on these benchmarks has been becoming a key measure of progress for action recognition models. (IV) \textit{Transfer learning:} due to the computational expense of training CNNs from scratch and the relatively smaller size of video datasets compared to image datasets, transfer learning became a popular strategy. Networks pre-trained on large image datasets like ImageNet \citep{deng2009imagenet} were fine-tuned on video frames for action recognition tasks. (V) \textit{Computational constraints:} despite the promise of CNNs, computational constraints were a significant challenge. Training deep networks required significant GPU power, and processing video data with CNNs was resource-intensive. This limited the complexity of the models that could be trained and the size of the datasets that could be used.

\subsubsection{Two-Stream CNNs}
\cite{simonyan2014two} presented an innovative approach to recognize actions in video sequences by using a two-stream CNN architecture. This approach divides the task into two distinct problems: recognizing spatial features from single frames and capturing temporal features across frames. The spatial stream CNN processes static visual information, while the temporal stream CNN handles motion by analyzing optical flow. The model was tested on benchmark datasets like UCF-101 and HMDB-51, where it achieved state-of-the-art results, showcasing the effectiveness of this two-stream method. The novelty of this work lies in the separation of motion and appearance features, which allows for more specialized networks that can better capture the complexities of video-based action recognition. The success of this model has made a significant impact on the field, influencing many future research directions in video understanding. Consequently, numerous methods have been proposed to enhance the the two-stream model \citep{wang2015towards, feichtenhofer2016convolutional,wang2016temporal,  peng2018two, wang2017spatiotemporal}.

In 2016, building on the the two-stream CNN, \cite{feichtenhofer2016convolutional} focused on improving the two-stream CNN by exploring various fusion strategies for combining spatial and temporal streams, resulting in better performance on the UCF-101 and HMDB-51 datasets. By enhancing fusion techniques, this work addressed the limitations of the initial two-stream model, leading to more effective integration of spatial and temporal information. \cite{wang2016temporal} introduced temporal segment networks (TSN). This work aimed to capture long-range temporal structures for action recognition, achieving significant improvements on the UCF-101 and HMDB-51 datasets by dividing videos into segments for comprehensive analysis. The introduction of TSN extended the temporal analysis capabilities of the two-stream CNN, enabling the capture of long-range dependencies.

In 2017, derived from the two-stream CNN, \cite{cosmin2017spatio} proposed a three-stream method by using spatio-temporal vectors, with locally max-pooled features to enhance performance. Tested on the UCF-101 and HMDB-51 datasets, the approach demonstrated improved recognition accuracy by efficiently capturing spatio-temporal dynamics. In 2018, the efficient convolutional network for online video understanding (ECO) was introduced by  \cite{zolfaghari2018eco}, combining the two-stream CNN approach with lightweight 3D CNNs, and focusing on efficiency and real-time processing, with high efficiency and competitive accuracy demonstrated on the Kinetics and UCF-101 datasets.

\cite{feichtenhofer2019slowfast} introduced the SlowFast network which processes video data at varying frame rates to capture both spatial semantics and motion dynamics, achieving state-of-the-art results on the Kinetics-400 and Charades datasets. By introducing different temporal resolutions, this work innovated on the two-stream concept, capturing fine and coarse temporal details. \cite{wang2018temporal} expanded on their previous work with TSN, developing a multi-stream approach that incorporated RGB, optical flow, and warped optical flow streams to model long-range temporal structures more effectively. This approach achieved state-of-the-art results by capturing both spatial and temporal information across various time scales. In 2021, temporal difference networks (TDN) were introduced by \cite{wang2021tdn}, leveraging the multi-stream CNN with a focus on capturing motion dynamics efficiently. Using the UCF-101 and HMDB-51 datasets, TDN achieved notable improvements by effectively modeling temporal differences. By emphasizing temporal differences, this work advanced the ability of the two-stream CNN to capture motion dynamics more effectively. 

Table \ref{table:TwoStreamNetworks} presents the works discussed in this section that utilized two or more stream CNNs approaches.

\begin{longtable}{
  |>{\centering\arraybackslash}p{2cm} % Column 1 width and alignment
  |>{\centering\arraybackslash}p{2cm}   % Column 2 width and alignment
  |>{\centering\arraybackslash}p{2cm} % Column 3 width and alignment
  |>{\centering\arraybackslash}p{8cm} % Column 4 width and alignment
  |>{\centering\arraybackslash}p{8cm}| % Column 5 width and alignment
}
\caption{Two-stream CNN-based approaches in HAR.} \label{tab:long_custom} \\
\hline
\textbf{Paper}  & \textbf{Model} & \textbf{Dataset} & \textbf{Novelty} \\
\hline
\endfirsthead

\multicolumn{4}{c}%
{{\bfseries \tablename\ \thetable{} -- continued from previous page}} \\
\hline
\textbf{Paper}  & \textbf{Model} & \textbf{Dataset} & \textbf{Novelty} \\
\hline
\endhead

\hline \multicolumn{5}{|r|}{{Continued on next page}} \\ \hline
\endfoot

\hline 
\endlastfoot

% Your table content here
\cite{simonyan2014two} & Two-stream CNN &  UCF-101, HMDB-51  & Introduced the two-stream architecture separating spatial and temporal streams for effective action recognition.                
  
\\ \midrule 
  
\cite{feichtenhofer2016convolutional} & Two-stream CNN & UCF-101, HMDB-51 & Explored various fusion strategies to combine spatial and temporal streams, and improved performance.   
\\ \midrule
\cite{wang2016temporal}  & Two-stream CNN + TSN & UCF-101, HMDB-51 & Introduced TSN to capture long-range temporal structures by dividing videos into segments.   
 
\\ \midrule
\cite{cosmin2017spatio} & Three-Stream CNN & UCF-101, HMDB-51 & Proposed a three-stream method using spatio-temporal vectors with locally max-pooled features for enhanced performance.
\\ \midrule
\cite{zolfaghari2018eco} & Two-stream CNN + 3D CNN & Kinetics, UCF-101 & Combined the two-stream CNN with lightweight 3D CNNs for efficient real-time processing.   
\\ \midrule
\cite{feichtenhofer2019slowfast}  & Two-stream CNN + SlowFast & Kinetics-400, Charades & Introduced SlowFast networks processing video data at varying frame rates to capture both spatial and motion dynamics.
\\ \midrule
\cite{wang2018temporal} &CNN-RNN, (Multi-stream TSN) & UCF101, HMDB51 & Expanded on TSN by developing a multi-stream approach that incorporated RGB, optical flow, and warped optical flow streams to model long-range temporal structures more effectively.
\\ \midrule
\cite{wang2021tdn} & Multi-stream CNN + TDN & Something-Something V1 and V2& Introduced TDN focusing on capturing motion dynamics efficiently.
% Repeat rows as needed
\label{table:TwoStreamNetworks}
\end{longtable}

\subsubsection{3D CNN-Based Approaches}

The foundational work conducted by \cite{ji20123d} introduced 3D CNNs for HAR, demonstrating their effectiveness in capturing spatio-temporal features on the KTH and UCF-101 datasets and outperforming traditional 2D CNNs. The work paved the way for further research on enhancing 3D convolutional models. \cite{tran2015learning} introduced C3D, a generic 3D CNN for spatio-temporal feature learning, achieving state-of-the-art performance on the Sports-1M and UCF-101 datasets and highlighting the scalability and effectiveness of 3D convolutions. Building on the work  by \cite{ji20123d}, C3D demonstrated the potential of 3D CNNs across diverse datasets, influencing subsequent research in 3D CNNs. \cite{varol2017long} introduced long-term temporal convolutions to capture extended motion patterns. This work improved the accuracy on the UCF-101 and HMDB-51 datasets and emphasized the importance of long-term motion information. Moreover, this study extended the temporal scope of 3D CNNs, highlighting the need for capturing long-term motion for accurate action recognition. In the same year, \cite{qiu2017learning} proposed pseudo-3D residual networks (P3D), which combined 2D and 3D convolutions to balance the accuracy and computational complexity. This work achieved competitive performance on the Kinetics and UCF-101 datasets. Moreover, P3D networks offered a more efficient approach by blending 2D and 3D convolutions, further refining the capabilities of 3D CNNs. Additionally, \cite{carreira2017quo} introduced I3D by inflating 2D convolutions to 3D, achieving significant improvements on the Kinetics dataset by leveraging ImageNet pre-training, thereby setting new performance benchmarks. I3D bridged the gap between 2D and 3D CNNs, demonstrating the benefits of transfer learning in 3D convolutional models.

\cite{hara2018can} evaluated the scalability of 3D CNNs with increased data and model sizes, demonstrating that deeper 3D CNNs can achieve better performance on the Kinetics and UCF-101 datasets, paralleling the success of 2D CNNs on ImageNet. This study emphasized the need for larger datasets and deeper models in 3D convolutional research, highlighting the potential of 3D CNNs to retrace the historical success of 2D CNNs. Building on these insights, \cite{diba2017temporal} introduced a new temporal 3D ConvNet architecture with enhanced transfer learning capabilities, demonstrating superior performance on the UCF-101 and HMDB-51 datasets through architectural innovations and effective transfer learning. This work underscored the importance of architectural innovation and transfer learning, pushing the boundaries of 3D CNN performance and further advancing the field of action recognition. \cite{tran2018closer} further contributed by conducting a comprehensive analysis of spatio-temporal convolutions, highlighting the benefits of factorizing 3D convolutions into separate spatial and temporal components, achieving state-of-the-art results on the Kinetics and UCF-101 datasets. This dissection provided insights that informed subsequent model designs and optimizations. In the same year, \cite{xie2018rethinking} explored the trade-offs between speed and accuracy in spatio-temporal feature learning, proposing efficient 3D CNN variants that balance computational cost and recognition performance on the Kinetics and UCF-101 datasets. Their work highlighted the practical considerations of deploying 3D CNNs, emphasizing the need to balance speed and accuracy, thereby refining the approach to spatio-temporal feature learning. Additionally, \cite{wang2018non} introduced non-local neural networks to capture long-range dependencies, demonstrating that non-local operations significantly improve the modeling of complex temporal relationships and enhance action recognition performance on the Kinetics and Something-Something datasets. By integrating non-local operations, this study advanced the ability of 3D CNNs to capture complex temporal patterns, further pushing the boundaries of spatio-temporal modeling.

\cite{feichtenhofer2019slowfast} introduced SlowFast Networks, a novel approach that processes video at different frame rates to capture both slow and fast motion dynamics, and achieved state-of-the-art results on the Kinetics-400 and Charades datasets. This innovation highlighted the importance of capturing varied motion dynamics for improved video recognition. In the same year, \cite{tran2019video} presented channel-separated convolutional networks (CSN), which reduced computational complexity by separating convolutions by channel, demonstrating efficiency without sacrificing accuracy on the Kinetics and Sports-1M datasets. This approach contributed to the development of more computationally feasible models. Concurrently, \cite{ghadiyaram2019large} leveraged large-scale weakly-supervised pre-training on video data, significantly boosting performance on the IG-65M and Kinetics datasets and underscoring the potential of massive datasets in enhancing 3D CNN capabilities. Additionally, \cite{kopuklu2019resource} proposed resource-efficient 3D CNNs using depthwise separable convolutions and achieved competitive accuracy with significantly reduced computational requirements on the Kinetics-400 and UCF-101 datasets. This work emphasized the importance of optimizing 3D CNNs for computational efficiency, further advancing the field of action recognition.

\cite{feichtenhofer2020x3d} proposed X3D, a family of efficient video models by expanding architectures along multiple axes. It achieved state-of-the-art performance with reduced model complexity on the Kinetics-400 and Charades datasets. X3D highlighted the significance of model efficiency in balancing performance and computational demands. In the same year, \cite{li2020spatio} introduced an efficient 3D CNN with a temporal attention mechanism and achieved high accuracy with efficient computation by focusing on salient temporal features on the Kinetics-400 and UCF-101 datasets. This work demonstrated the potential of selectively focusing on important temporal features to enhance the efficiency and accuracy of 3D CNNs, further advancing the field of action recognition. 

Table \ref{table:3DCNNBased} presents the works discussed in this section that utilized 3D CNN approaches.

\begin{longtable}{
  |>{\centering\arraybackslash}p{2cm} % Column 1 width and alignment
  |>{\centering\arraybackslash}p{2cm}   % Column 2 width and alignment
  |>{\centering\arraybackslash}p{2cm} % Column 3 width and alignment
  |>{\centering\arraybackslash}p{8cm} % Column 4 width and alignment
  |>{\centering\arraybackslash}p{8cm}| % Column 5 width and alignment
}
\caption{3D CNN-based approaches in HAR.} \label{tab:long_custom} \\
\hline
\textbf{Paper} & \textbf{Model} & \textbf{Dataset} & \textbf{Novelty} \\
\hline
\endfirsthead

\multicolumn{4}{c}%
{{\bfseries \tablename\ \thetable{} -- continued from previous page}} \\
\hline
\textbf{Paper}  & \textbf{Model} & \textbf{Dataset} & \textbf{Novelty} \\
\hline
\endhead

\hline \multicolumn{4}{|r|}{{Continued on next page}} \\ \hline
\endfoot

\hline 
\endlastfoot

% Your table content here
\cite{ji20123d}  & 3D CNN &  UCF-101, HMDB-51  &  Introduced 3D CNNs for HAR, effectively capturing spatio-temporal features and outperforming 2D CNNs.                
\\ \midrule

\cite{tran2015learning} & 3D CNN &  Sports-1M, UCF-101  & Introduced C3D, a generic 3D CNN for spatio-temporal feature learning, and achieved state-of-the-art performance.                 
\\ \midrule

\cite{varol2017long}  & 3D CNN &  UCF-101, HMDB-51  &   Introduced long-term temporal convolutions to capture extended motion patterns, and improved accuracy.               
\\ \midrule

\cite{qiu2017learning}& 3D CNN &  Kinetics, UCF-101 &   Proposed P3D networks combining 2D and 3D convolutions, balancing accuracy and computational complexity.               
\\ \midrule

\cite{carreira2017quo}& 3D CNN &  Kinetics  &  Introduced I3D by inflating 2D convolutions to 3D, leveraging ImageNet pre-training for significant improvements.                
\\ \midrule
\cite{hara2018can}& 3D CNN &  Kinetics, UCF-101  &  Evaluated the scalability of 3D CNNs with increased data and model sizes, and showed parallels to 2D CNN success.                
\\ \midrule

\cite{diba2017temporal} & 3D CNN &  UCF-101, HMDB-51  & Introduced a new temporal 3D ConvNet architecture with enhanced transfer learning capabilities.                 
\\ \midrule

\cite{tran2018closer} & 3D CNN & Kinetics, UCF-101  &  Conducted a comprehensive analysis of spatio-temporal convolutions, and highlighted the benefits of factorizing 3D convolutions.                
\\ \midrule

\cite{xie2018rethinking}& 3D CNN & Kinetics, UCF-101  &  Explored speed-accuracy trade-offs in spatio-temporal feature learning, and proposed efficient 3D CNN variants.                
\\ \midrule

\cite{wang2018non}& 3D CNN & Kinetics, Something-Something  &  Introduced non-local operations to capture long-range dependencies, and improved modeling of complex temporal relationships.                
\\ \midrule

\cite{feichtenhofer2019slowfast}& 3D CNN & Kinetics-400, Charades  &  Proposed SlowFast networks to process video at different frame rates, capturing both slow and fast motion dynamics.                
\\ \midrule
\cite{tran2019video}& 3D CNN & Kinetics, Sports-1M  & Introduced CSN to reduce computational complexity without sacrificing accuracy.                 
\\ \midrule

\cite{ghadiyaram2019large}& 3D CNN & IG-65M, Kinetics  &   Leveraged large-scale weakly-supervised pre-training on video data, and significantly boosted performance.               
\\ \midrule

\cite{kopuklu2019resource} & 3D CNN & Kinetics-400, UCF-101  &   Proposed resource-efficient 3D CNNs using depthwise separable convolutions, and achieved competitive accuracy with reduced computational requirements.               
\\ \midrule

\cite{feichtenhofer2020x3d}& 3D CNN & Kinetics-400, Charades  &  Proposed X3D, a family of efficient video models by expanding architectures along multiple axes.                
\\ \midrule

\cite{li2020spatio}& 3D CNN & Kinetics-400, UCF-101  &  Introduced a temporal attention mechanism to enhance efficiency and accuracy in 3D CNNs.    
% Repeat rows as needed
\label{table:3DCNNBased}
\end{longtable}

\subsubsection{CNN-RNN-Based Approaches}

The integration of CNNs and RNNs for HAR was significantly advanced by the work of \cite{donahue2015long}, who introduced long-term recurrent convolutional networks (LRCN). This approach effectively combined the spatial feature extraction capabilities of CNNs with the temporal dynamics modeling of LSTMs, demonstrating substantial improvements in action recognition tasks on datasets like UCF-101 and HMDB-51. Building on this foundation, \cite{yue2015beyond} extended the application of deep networks to video classification by integrating deep CNNs with LSTMs to handle longer video sequences. Their method, tested on the Sports-1M and UCF-101 datasets, highlighted the importance of capturing extended temporal dependencies for improved performance in complex video classification tasks. Further pushing the boundaries, \cite{srivastava2015unsupervised} explored unsupervised learning of video representations using LSTMs. By leveraging LSTMs to learn spatio-temporal features without labeled data, their approach demonstrated effective video representation learning on the UCF-101 dataset, showcasing the versatility and potential of CNN-RNN architectures in both supervised and unsupervised learning scenarios for HAR.

The development of CNN-RNN architectures for HAR saw significant advancements in 2016.  \cite{wu2015modeling} proposed a hybrid deep learning framework that modeled spatial-temporal clues by combining CNNs for spatial feature extraction with RNNs for temporal sequence modeling. Their approach, tested on the UCF-101 and HMDB-51 datasets, demonstrated substantial improvements in video classification accuracy. Additionally, \cite{li2016online}  expanded the application of CNN-RNN architectures to real-time scenarios with their approach for online human action detection using joint classification-regression RNNs. Combining CNNs for spatial features and RNNs for temporal dynamics, their method, tested on the J-HMDB and UCF-101 datasets, achieved notable improvements in accuracy and efficiency, showcasing the practicality of CNN-RNN models in real-time action detection. 

Building on these advancements, 2017 and 2018 witnessed further refinements and innovations in CNN-RNN architectures for HAR. \cite{li2018videolstm} introduced VideoLSTM, integrating convolutions, attention mechanisms and optical flow within a recurrent framework, and demonstrating improved performance on the UCF101 and HMDB51 datasets. \cite{carreira2017quo} made a significant contribution with the two-stream Inflated 3D ConvNet (I3D), which inflated 2D CNN architectures into 3D and combined them with RNNs for temporal modeling. The model was evaluated on the Kinetics dataset, as well as UCF101 and HMDB51. \cite{ullah2017action} proposed a novel architecture combining CNNs with bi-directional LSTMs, effectively utilizing both spatial and temporal information from video sequences and showing superior performance on the UCF-101 and HMDB-51 datasets. In 2020, in the realm of human activity recognition using sensor data, \cite{xia2020lstm} proposed an LSTM-CNN architecture that effectively captured both temporal dependencies and local feature patterns, showing improved accuracy on the WISDM, UCI HAR, and OPPORTUNITY datasets. Similarly, \cite{mutegeki2020cnn} developed a CNN-LSTM approach for smartphone sensor-based activity recognition, demonstrating high accuracy on the UCI HAR dataset and further validating the effectiveness of combining CNNs and RNNs for processing time-series data in activity recognition tasks.

Recent advancements in HAR have leveraged sophisticated CNN-RNN architectures to enhance performance and reduce computational complexity. \cite{muhammad2021human} introduced an attention-based LSTM network combined with dilated CNN features, and significantly improved the recognition accuracy on the UCF-101 and HMDB-51 datasets by capturing essential spatial features through dilated convolutions and temporal patterns with attention mechanisms. Building on this, \cite{malik2023cascading} focused on multiview HAR; utilizing a CNN-LSTM architecture to cascade pose features, they achieved high accuracy (94.4\% on the MCAD dataset and 91.67\% on the IXMAS dataset) while reducing the computational load by targeting pose data rather than entire images. 

Table \ref{table:CNNRNNbasedApproaches} presents the works discussed in this section that utilized CNN-RNN approaches.

\begin{longtable}{
  |>{\centering\arraybackslash}p{2cm} % Column 1 width and alignment
  |>{\centering\arraybackslash}p{2cm}   % Column 2 width and alignment
  |>{\centering\arraybackslash}p{2cm} % Column 3 width and alignment
  |>{\centering\arraybackslash}p{8cm} % Column 4 width and alignment
  |>{\centering\arraybackslash}p{8cm}| % Column 5 width and alignment
}
\caption{CNN-RNN-based approaches in HAR.} \label{tab:long_custom} \\
\hline
\textbf{Paper} & \textbf{Model} & \textbf{Dataset} & \textbf{Novelty} \\
\hline
\endfirsthead

\multicolumn{4}{c}%
{{\bfseries \tablename\ \thetable{} -- continued from previous page}} \\
\hline
\textbf{Paper} &  \textbf{Model} & \textbf{Dataset} & \textbf{Novelty} \\
\hline
\endhead

\hline \multicolumn{4}{|r|}{{Continued on next page}} \\ \hline
\endfoot

\hline 
\endlastfoot

\cite{donahue2015long} &CNN-RNN, (LRCN)&UCF-101, HMDB-51 & Combined CNNs for spatial feature extraction with LSTMs for temporal dynamics.

\\ \midrule

\cite{yue2015beyond}&CNN-RNN &Sports-1M, UCF-101& Integrated deep CNNs with LSTMs to handle longer video sequences, capturing extended temporal dependencies.

\\ \midrule
\cite{srivastava2015unsupervised}&CNN-RNN, (Unsupervised LSTM)&UCF-101& Explored unsupervised learning of video representations using LSTMs, leveraging spatiotemporal features.

\\ \midrule
\cite{wu2015modeling} &CNN-RNN & UCF-101, HMDB-51 & Modeled spatial-temporal clues by combining CNNs for spatial features with RNNs for temporal sequence modeling.

\\ \midrule
\cite{li2016online}&CNN-RNN&J-HMDB, UCF-101& Applied CNN-RNN architectures to real-time scenarios for online human action detection.

\\ \midrule
\cite{li2018videolstm} &CNN-RNN (VideoLSTM)&UCF-101, HMDB-51&Integrated convolutions, attention mechanisms, and optical flow within a recurrent framework.

\\ \midrule
\cite{carreira2017quo}&3D CNN-RNN&Kinetics, UCF101, HMDB51&Inflated 2D CNN architectures into 3D, and combined them with RNNs for temporal modeling.
\\ \midrule
\cite{ullah2017action}&CNN-RNN, (CNN-BiLSTM)&UCF101, HMDB51&Combined CNNs with bi-directional LSTMs to utilize both spatial and temporal information.

\\ \midrule
\cite{xia2020lstm}&CNN-RNN &WISDM, UCI, OPPORTUNITY&Captured both temporal dependencies and local feature patterns for human activity recognition using sensor data.
\\ \midrule
\cite{mutegeki2020cnn}&CNN-RNN &UCI& Developed a CNN-LSTM approach for smartphone sensor-based activity recognition, and demonstrated high accuracy.
\\ \midrule
\cite{muhammad2021human}&CNN-RNN, (CNN-Attention-LSTM)&UCF-101, HMDB-51  &Improved recognition accuracy with attention-based LSTM network combined with dilated CNN features.
\\ \midrule
\cite{malik2023cascading}&CNN-RNN &MCAD, IXMAS&Achieved high accuracy in multiview HAR  by cascading pose features using a CNN-LSTM architecture.
% Repeat rows as needed
\label{table:CNNRNNbasedApproaches}
\end{longtable}

\subsection{ViT-Based Approaches in HAR}
In 2020, the ViT was conceptualized and introduced in the academic domain through the paper authored by \cite{dosovitskiy2020image}. The ViT marked a paradigm shift in still image recognition methodologies, applying the Transformer model, predominantly known for its success in NLP, to the realm of computer vision. The application of ViTs in action recognition, a more specific and complex task within the field of computer vision, followed the initial introduction of ViT. Specifically, in 2021 and beyond, subsequent research and publications have explored and expanded the use of ViTs for action recognition tasks, demonstrating their efficacy in capturing spatial-temporal features within video data. They employ attention mechanisms to minimize redundant information and to model interactions over long distances in both space and time \citep{koot2021evaluating}. The adaptation of ViT to action recognition signifies the model's versatility and its potential for broader applications in computer vision beyond static image analysis.

Recent advancements in action recognition have seen a significant shift towards ViT, highlighting their efficacy in video understanding tasks. \cite{arnab2021vivit} introduced ViViT, extending the vision Transformer architecture to handle video sequences. They demonstrated its potential on datasets like Kinetics-400 and Something-Something-V2, marking a substantial improvement in video action recognition capabilities. Building on this, \cite{bertasius2021space} proposed a space-time Transformer that models temporal information innovatively,  and achieved competitive results on similar datasets. The efficiency of multiscale ViTs was further illustrated by \cite{fan2021multiscale}, who showed that such architectures could effectively capture fine-grained video details and enhance classification performance on comprehensive video datasets. Moreover, \cite{liu2022video} presented the Swin Transformer, utilizing a shifted window mechanism to model long-range dependencies more efficiently, and leading to significant improvements in action recognition accuracy. Together, these works underscore the transformative impact of ViTs in advancing the field of HAR. Additionally, \cite{wang2021actionclip} introduced ActionCLIP, leveraging the CLIP model for enhanced video action recognition on multiple standard video datasets, including Kinetics-400 and HMDB-51. This novel approach integrated visual and linguistic representations.

\cite{chen2022mm} introduced Mm-ViT, a multi-modal video Transformer designed for compressed video action recognition, and demonstrated high performance by leveraging multi-modal inputs on compressed video datasets such as HACS and UCF101. \cite{sharir2021image} explored the extension of ViT to video data, showing its potential in capturing temporal dynamics effectively across several standard video datasets including Kinetics-400 and HMDB-51. Furthermore, \cite{xing2023svformer} developed SVFormer, a semi-supervised video Transformer that leverages both labeled and unlabeled data to bridge the gap between supervised and unsupervised learning, and achieved significant improvements in action recognition tasks on various standard HAR datasets such as Kinetics-400 and UCF101. Together, these works underscore the transformative impact of ViTs in advancing the field of HAR. 

Table \ref{table:TransformerbasedApproaches1} presents the works discussed in this section that utilized ViTs.

\begin{longtable}{
  |>{\centering\arraybackslash}p{2cm} % Column 1 width and alignment
  |>{\centering\arraybackslash}p{2cm}   % Column 2 width and alignment
  |>{\centering\arraybackslash}p{2cm} % Column 3 width and alignment
  |>{\centering\arraybackslash}p{8cm} % Column 4 width and alignment
  |>{\centering\arraybackslash}p{8cm}| % Column 5 width and alignment
}
\caption{ViT-based approaches in HARs.} \label{tab:long_custom} \\
\hline
\textbf{Paper}  & \textbf{Model} & \textbf{Dataset} & \textbf{Novelty} \\
\hline
\endfirsthead

\multicolumn{4}{c}%
{{\bfseries \tablename\ \thetable{} -- continued from previous page}} \\
\hline
\textbf{Paper}  & \textbf{Model} & \textbf{Dataset} & \textbf{Novelty} \\
\hline
\endhead

\hline \multicolumn{4}{|r|}{{Continued on next page}} \\ \hline
\endfoot

\hline 
\endlastfoot

\cite{arnab2021vivit} &ViViT&Kinetics-400, Something-Something-V2&Extended ViT to video sequences.

\\ \midrule

\cite{bertasius2021space}&Space-Time Transformer&Kinetics-400&Innovative temporal information modeling.

\\ \midrule
\cite{fan2021multiscale}&Multiscale ViT&Kinetics-400, Something-Something-V2&Efficient capture of fine-grained video details.

\\ \midrule

\cite{liu2022video}&Swin Transformer&Kinetics-400, Something-Something-V2&Shifted window mechanism for long-range dependency modeling.
\\ \midrule
\cite{wang2021actionclip}&ActionCLIP&Kinetics-400, HMDB-51&Leveraged CLIP for enhanced video action recognition.

\\ \midrule
\cite{chen2022mm} &Mm-ViT&HACS, UCF101&Multi-modal inputs for compressed video action recognition.

\\ \midrule
\cite{sharir2021image}&ViT&Kinetics-400, HMDB-51&Applied ViT to video data.

\\ \midrule
\cite{xing2023svformer}&SVFormer&Kinetics-400, UCF101&Semi-supervised learning for action recognition.
\label{table:TransformerbasedApproaches1}
\end{longtable}

\subsection{CNN-ViT Hybrid Architectures}
The integration of ViTs with CNNs has significantly advanced HAR tasks. \cite{zhang2021twostream} proposed a two-stream hybrid CNN-Transformer network (THCT-Net), which demonstrated enhanced generalization ability and convergence speed on the NTU RGB+D dataset by combining CNNs for low-level context sensitivity and Transformers for capturing global information. Following this, \cite{jegham2022multiview} applied a similar hybrid model to driver action recognition, leveraging multi-view data to achieve high accuracy through the integration of CNNs for spatial feature extraction and Transformers for temporal dependencies. \cite{kalfaoglu2022human} extended this approach by integrating 3D CNNs with Transformers for late temporal modeling, and achieved substantial improvements in action recognition accuracy on the HMDB-51 and UCF101 datasets. Moreover, \cite{yu2023swinfusion} proposed Swin-Fusion, which combines Swin Transformers with CNN-based feature fusion to achieve state-of-the-art performance on datasets like Kinetics-400 and Something-Something-V2, demonstrating the robustness and superior performance of hybrid models in HAR tasks.

\cite{djenouri2022hybrid} proposed a hybrid visual Transformer model that integrates CNNs and Transformers for efficient and accurate human activity recognition. They demonstrated its capability on datasets like Kinetics-400 and UCF101, and showed that the hybrid approach leverages the local feature extraction of CNNs with the global context modeling of Transformers. Following this, \cite{surek2023videobased} provided a comprehensive review of deep learning approaches for video-based human activity recognition, emphasizing the potential of hybrid models. This review underscored the effectiveness of such hybrid models in capturing both spatial and temporal features from video data, and evaluated on various human activity datasets including NTU RGB+D and UTD-MHAD. \cite{ahmadabadi2023distilling} explored the use of knowledge distillation techniques to enhance the performance of hybrid CNN-Transformer models. Their approach was validated on datasets such as HMDB-51 and Kinetics-400, showing significant improvements in HAR by effectively transferring knowledge from complex models to more efficient ones. Together, these works highlight the evolving landscape of hybrid models in human activity recognition, showcasing their robustness and efficiency in handling complex video data. 

Table \ref{table:3DCNNBased} presents the works discussed in this section that utilized CNN-ViT approaches.

\begin{longtable}{
  |>{\centering\arraybackslash}p{2cm} % Column 1 width and alignment
  |>{\centering\arraybackslash}p{2cm}   % Column 2 width and alignment
  |>{\centering\arraybackslash}p{2cm} % Column 3 width and alignment
  |>{\centering\arraybackslash}p{8cm} % Column 4 width and alignment
  |>{\centering\arraybackslash}p{8cm}| % Column 5 width and alignment
}
\caption{ CNN-ViT hybrid approaches in HARs.} \label{tab:long_custom} \\
\hline
\textbf{Paper} & \textbf{Model} & \textbf{Datase} & \textbf{Novelty} \\
\hline
\endfirsthead

\multicolumn{4}{c}%
{{\bfseries \tablename\ \thetable{} -- continued from previous page}} \\
\hline
\textbf{Paper} & \textbf{Model} & \textbf{Datase} & \textbf{Novelty} \\
\hline
\endhead

\hline \multicolumn{4}{|r|}{{Continued on next page}} \\ \hline
\endfoot

\hline 
\endlastfoot

\cite{zhang2021twostream} &The two-stream hybrid CNN-Transformer network (THCT-Net)&NTU RGB+D&Combined CNNs and Transformers for improved generalization and convergence speed.
\\ \midrule
\cite{jegham2022multiview}&Multi-view vision Transformer&Custom driver action datasets & Leveraged multi-view data for spatial and temporal feature integration.
\\ \midrule
\cite{kalfaoglu2022human} &3D CNN-Transformer&HMDB-51, UCF101& Integrated 3D CNNs with Transformers for late temporal modeling.
\\ \midrule
\cite{yu2023swinfusion}&Swin-Fusion&Kinetics-400, Something-Something-V2 & Combined Swin Transformers with CNN-based feature fusion for state-of-the-art performance
\\ \midrule
\cite{djenouri2022hybrid} &Hybrid visual Transformer&Kinetics-400, UCF101 & Efficient and accurate human activity recognition leveraging strengths of CNNs and Transformers
\\ \midrule
\cite{surek2023videobased}&Various deep learning models including hybrid models&NTU RGB+D, UTD-MHAD& Comprehensive review highlighting the potential of hybrid models.
\\ \midrule
\cite{ahmadabadi2023distilling}&Hybrid CNN-Transformer&HMDB-51, Kinetics-400&Knowledge distillation from CNN-Transformer models for enhanced performance.
\label{table:HybridBasedApproaches1}
\end{longtable}

\subsection{Discussion}
In the field of HAR, the choice of models – whether CNN-based, ViT-based, or a hybrid of CNN and ViT – significantly influences the outcome and efficiency of the task. CNN-based models are particularly adept at extracting local features due to their convolutional nature \citep{lecun2015deep}, making them highly effective in pattern recognition within images and videos. Their computational efficiency is a boon for real-time applications \citep{howard2017mobilenets}, and their robustness to input variations is notable \citep{simonyan2014very}. However, CNNs often struggle with global contextual understanding \citep{szegedy2015going} and are prone to overfitting. Moreover, their ability to model long-range temporal dependencies \citep{karpathy2014large}, which is crucial in action recognition, is somewhat limited.

ViT-based models, in contrast, excel in capturing global dependencies \citep{carion2020end, dosovitskiy2020image}, thanks to their self-attention mechanism. This attribute makes them particularly suited for understanding complex actions that require a broader view beyond local features. ViTs are scalable with data, benefiting significantly from larger datasets, and are flexible in processing inputs of various sizes \citep{touvron2021training}. The adaptability in processing various input sizes is a byproduct of the patch-based approach and the global receptive field of the ViTs. However, these models are computationally more intensive and require substantial training data to achieve optimal performance \citep{khan2022Transformers}. Unlike CNNs, ViTs are not as efficient in extracting detailed local features, which can be a critical drawback in certain action recognition scenarios.

Hybrid models that combine CNNs and ViTs aim to harness the strengths of both architectures. They offer the local feature extraction capabilities of CNNs along with the global context awareness of ViTs, potentially providing a more balanced approach to action recognition. These models can be more efficient and versatile, adapting well to a range of tasks. However, this combination brings its own challenges, including increased architectural complexity, higher resource demands, and the need for careful tuning to balance the contributions of both CNN and ViT components. The choice among these models depends on the specific requirements of the action recognition task, such as the available computational resources, the nature and size of the dataset, and the types of actions that need to be recognized. 

For a summary of the advantages and disadvantages of these three architectural variations, see Table \ref{table:ACT}.

%\begin{comment}

\begin{table}[htbp]
\centering
\caption{Capability comparison between Transformer-based, CNN-based, and hybrid models in HARs.}

\begin{tabularx}{\textwidth}{X|c|c|c}
\toprule
\multicolumn{1}{c|}{\textbf{Criteria}} & \textbf{ViT-based} & \textbf{CNN-based} & \textbf{Hybrid Models} \\
\midrule
\midrule
\multicolumn{4}{c}{\textbf{Advantages}} \\
\midrule
Excel at capturing global dependencies & \checkmark &  & \checkmark \\
\midrule
Scalable with data & \checkmark &  & \checkmark \\
\midrule
Flexible in processing various input sizes & \checkmark &  & \checkmark \\
\midrule
Adept at extracting local features &  & \checkmark & \checkmark \\
\midrule
Computationally efficient &  & \checkmark &  \\
\midrule
Robust to input variations &  & \checkmark & \checkmark \\
\midrule
Efficient and versatile &  &  & \checkmark \\
\midrule
Adapts well to a range of tasks &  &  & \checkmark \\
\midrule
\multicolumn{4}{c}{\textbf{Disadvantages}} \\
\midrule
Computationally intensive & \checkmark &  & \checkmark \\
\midrule
Requires substantial training data & \checkmark &  & \checkmark \\
\midrule

Limited global contextual understanding &  & \checkmark &  \\
\midrule
Prone to overfitting &  & \checkmark &  \\
\midrule
Limited in modeling long-range dependencies &  & \checkmark &  \\
\midrule
Architectural complexity &  &  & \checkmark \\
\midrule
Higher resource demands &  &  & \checkmark \\
\midrule
Need for careful tuning &  &  & \checkmark \\
\midrule
Balancing contributions of both components can be challenging &  &  & \checkmark \\
\bottomrule
\end{tabularx}
\label{table:ACT}
\end{table}

%\end{comment}

%-----------------------------
\section{Proposed CNN-ViT Hybrid Architecture}
\label{sec:sec4}
\begin{figure}[H]

\includegraphics[scale=0.7]{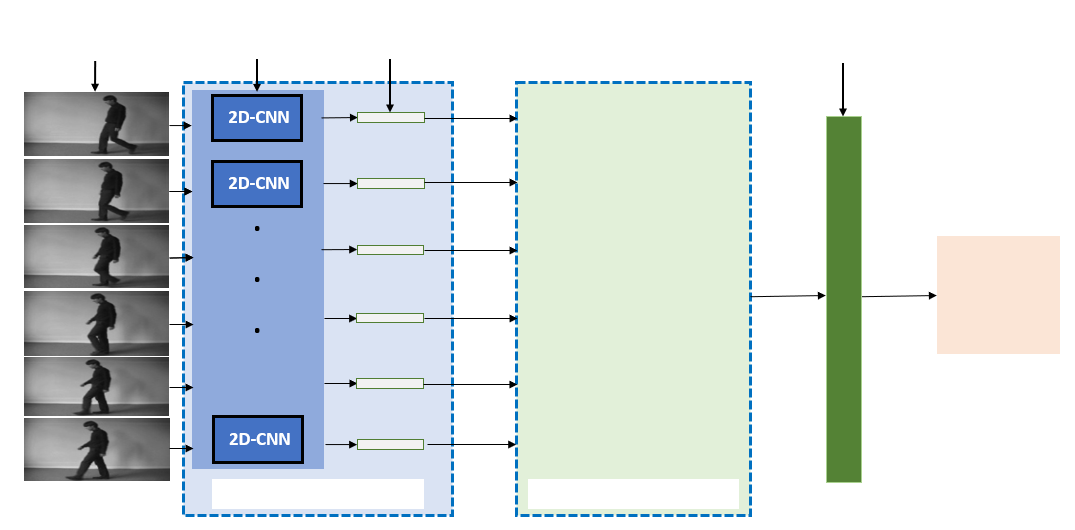}
\put(-195,95){\scriptsize ViT}
\put(-45,95){\scriptsize Class}
\put(-229,12){\scriptsize Temporal Component}
\put(-355,12){\scriptsize Spatial Component}
\put(-440,205){\scriptsize Video Frames}
\put(-380,210){\scriptsize TimeDistributed}
\put(-360,200){\scriptsize Layer}
\put(-303,210){\scriptsize Spatial }
\put(-305,200){\scriptsize  Features}
\put(-110,210){\scriptsize Softmax }
\put(-105,200){\scriptsize  Layer}
\put(-303,177){\scriptsize $\vect{v}_1$ }
\put(-303,150){\scriptsize $\vect{v}_2$ }
\put(-303,135){\scriptsize $.$ }
\put(-303,130){\scriptsize $.$ }
\put(-303,125){\scriptsize $.$ }
\put(-303,93){\scriptsize $\vect{v}_{N-2}$ }
\put(-303,67){\scriptsize $\vect{v}_{N-1}$ }
\put(-303,40){\scriptsize $\vect{v}_N$ }
\put(-120,100){\scriptsize $\vect{z}$ }

\caption{The hybrid CNN-ViT architecture for HARs.}
\label{fig:HybridModel}
\end{figure}

%-----------------------------
In this section, we present our proposed CNN-ViT architecture for HAR, leveraging the benefits of both approaches described in previous sections, see Figure \ref{fig:HybridModel}. The architecture incorporates a TimeDistributed layer with a CNN backbone, followed by a ViT model to classify actions in video sequences.

\textit{Spatial component.}
Let $\mathcal{X}$ be a collection of $N$ frames, i.e., $\mathcal{X}=\{\boldsymbol{X}_i\}_{i=1}^N$.
The CNN backbone (i.e. MobileNet in \citealt{howard2017mobilenets}) in the TimeDistributed layer (see Figure \ref{fig:HybridModel}) processes the indifivual frames $\vect{X}_i$ and outputs the spatial features vector $\vect{v}_i = p_\theta(\vect{X}_i) \in \mathbb{R}^L$, where $p_\theta$ is the CNN model (e.g. MobileNet or VGG16) with parameters in $\theta$ wrapped by the TimeDistributed layer. 

%The collection of $n$ frames $\mathcal{X}$ is fed into the spatial component of our architecture in Figure \ref{fig:HybridModel}.
%The trained CNN $p_{\theta}$ is then used $n$ times to process $\mathcal{X}$ frame by frame and create a set of $n$ spatial feature vectors $\mathcal{V}=\{\boldsymbol{v}^i\}_{i=1}^n$, where $\boldsymbol{v}^i = p_\theta(\boldsymbol{X}^i)$.

\textit{Temporal component.}
In the proposed hybrid CNN-ViT model, ViT is engineered to process the sequence of the $N$ spatial features vectors, i.e., $\{\boldsymbol{v}_i\}_{i=1}^N$, where each $\boldsymbol{v}_i$ represents a distinct frame of the input video clip, see Figure \ref{fig:HybridModel}. Afterwards, the ViT block outputs a final representation $\vect{z}$, which is then fed into the softmax layer to classify the action in the video. In detail, the Transformer encoder is designed to process a sequence of vectors, each representing  one frame, and aggregate information into a single vector for classification.

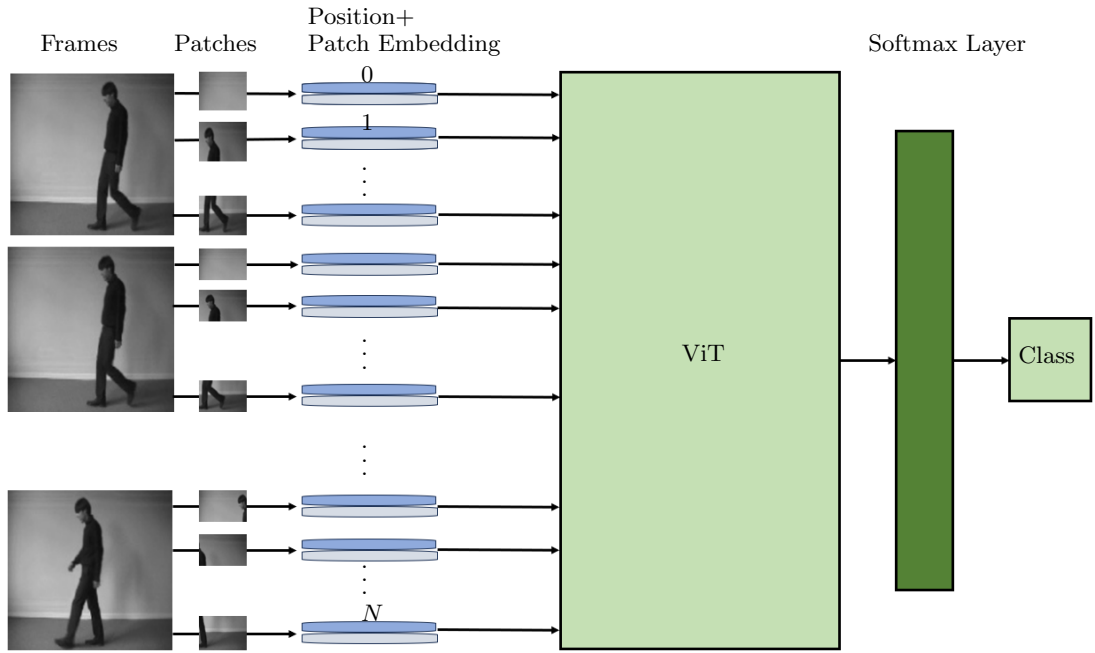
\begin{figure}[H]

\includegraphics[scale=0.55]{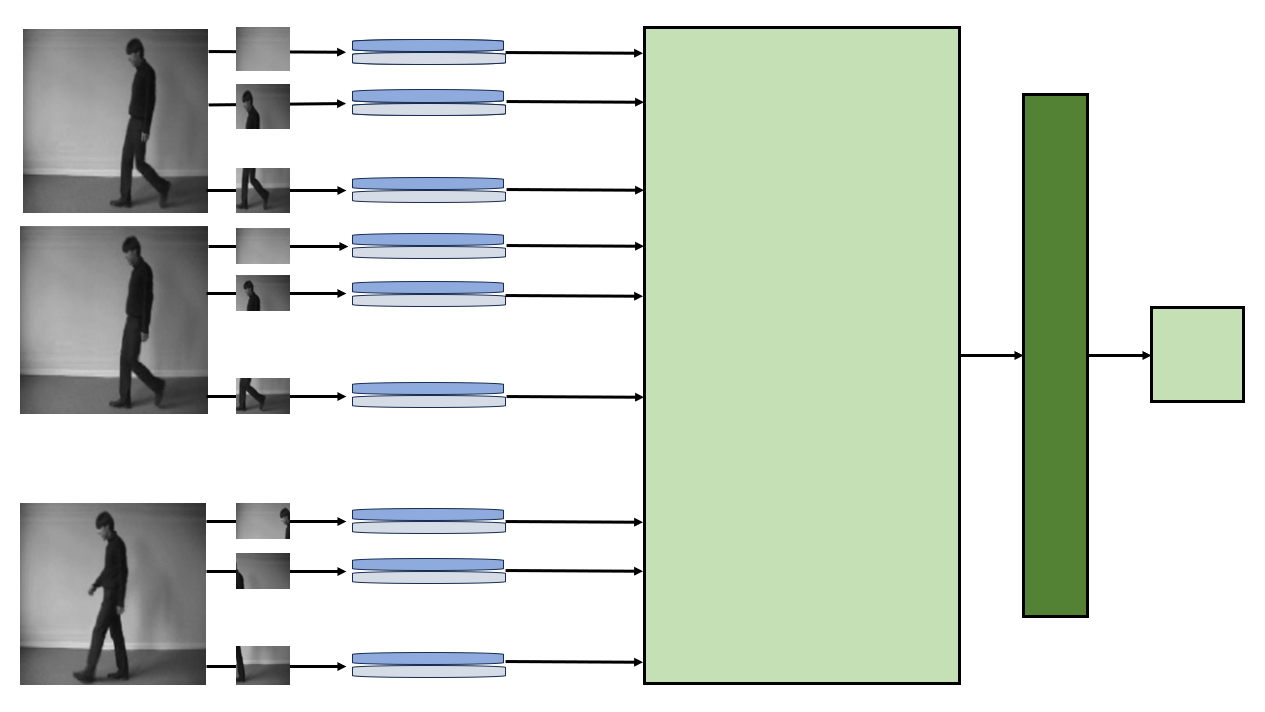}
\put(-160,118){\scriptsize ViT}
\put(-34,117){\scriptsize Class}
\put(-300,245){\scriptsize Position+}
\put(-300,235){\scriptsize Patch Embedding}
\put(-350,235){\scriptsize Patches}
\put(-400,235){\scriptsize Frames}
\put(-90,235){\scriptsize Softmax Layer}
\put(-280,223){\scriptsize $0$}
\put(-280,205){\scriptsize $1$}
\put(-280,20){\scriptsize $N$}
\put(-280,30){\scriptsize $.$}
\put(-280,35){\scriptsize $.$}
\put(-280,40){\scriptsize $.$}
\put(-280,75){\scriptsize $.$}
\put(-280,80){\scriptsize $.$}
\put(-280,85){\scriptsize $.$}
\put(-280,115){\scriptsize $.$}
\put(-280,120){\scriptsize $.$}
\put(-280,125){\scriptsize $.$}
\put(-280,180){\scriptsize $.$}
\put(-280,185){\scriptsize $.$}
\put(-280,190){\scriptsize $.$}

\caption{The ViT-only architecture for HARs.}
\label{fig:ViTModel}
\end{figure}

In the proposed ViT-only model in Figure \ref{fig:ViTModel} for the purpose of comparison, each vector represents a distinct patch. These vectors are first linearly projected into a high-dimensional space, facilitating the model's ability to learn complex patterns within the data. To ensure the model captures the sequential nature of the input, positional encodings are added to these embeddings. The core of the ViT consists of two layers, each comprising a multi-head self-attention mechanism and a feed-forward network. The self-attention mechanism allows the model to weigh the importance of different patches relative to each other, while the feed-forward network, utilizing an exponential linear unit (ELU) activation function, processes each position independently to capture global context. The ViT is designed to aggregate the information from all vectors and positional encodings into a single [CLS] token, which is prepended to the input sequence. The output vector associated with this [CLS] token, after propagation through the Transformer layers, serves as a comprehensive representation of the entire input, suitable for downstream classification tasks.

\subsection{Experiments}
The goal of the presented experiments is not necessarily to produce a model that outperforms the state-of-the-art models in the HAR field. Rather, the aim is to conduct a comparison among the CNN, ViT-only, and hybrid models to give further insights. 

The Royal Institute of Technology in 2004 unveiled the KTH dataset, a significant and publicly accessible dataset for action recognition \citep{schuldt2004recognizing}. The KTH dataset was chosen here for its balanced representation of spatial and temporal features. Renowned as a benchmark dataset, it encompasses six types of actions: walking, jogging, running, boxing, hand-waving, and hand-clapping. The dataset features performances by 25 different individuals, introducing a diversity in execution. Additionally, the environment for each participant's actions was deliberately altered, including settings such as outdoors, outdoors with scale changes, outdoors with clothing variations, and indoors. The KTH dataset comprises 2,391 video sequences, all recorded at 25 frames per second using a stationary camera against uniform backgrounds. 

Six experiments were conducted, with  each of the aforementioned models trained on three different lengths of frame sequences. Care was taken to avoid pre-training in order to ensure the neutrality of the results. The TransNet model by \cite{alomar2023transnet} was adopted to represent the CNN model, and the ViT model was depicted in Figure \ref{fig:ViTModel}. For the spatial component of the hybrid model, we employed the spatial component of TransNet; and for the temporal component, we employed the same ViT model that we used in the ViT-only model.
We constructed our model utilizing Python 3.6, incorporating the Keras deep learning framework, OpenCV for image processing, matplotlib, and the scikit-learn library. The training and test were performed on a computer equipped with an Intel Core i7 processor, an NVidia RTX 2070 graphics card, and 64GB of RAM.

%-------
{\subsubsection{Results and Discussion }}
%-------

\begin{table}[h]
\caption{Experimental results of different models on the KTH Dataset using three different context lengths. In particular, the hybrid model was trained without pre-training whereas Hybrid$_{\rm pre}$ is for the hybrid model pre-trained on ImageNet.} % Add a caption if needed
\label{table:ModelsComparision}
\centering
\begin{tabular}{c||c|c|c|c}
\hline 
\textbf{Context length} & \textbf{CNN-based} & \textbf{ViT-only} &  \textbf{Hybrid} & \textbf{Hybrid}$_{\rm pre}$ \\
\hline \hline
12 frames & 94.35 & 92.44 & 94.12& 96.34 \\ 
18 frames & 93.91 &92.82 & 94.56 & 97.13 \\ 
24 frames & 93.49 & 93.69 & 95.78 & \textbf{97.89} \\ \hline
\end{tabular}
\end{table}

Table \ref{table:ModelsComparision} presents the quantitative results of the three distinct models, i.e., CNN, ViT-only, and a hybrid model on the KTH dataset, focusing on three different context lengths, i.e., short (12 frames), medium (18 frames), and long (24 frames). The results from these experiments provide insightful revelations into the efficacy of each model under different temporal contexts. More details are given below.

The CNN model exhibited a decrease in accuracy as the frame length increased, recording 94.35\% for 12 frames, 93.91\% for 18 frames, and 93.49\% for 24 frames. This descending trend suggests that CNN may struggle with processing longer sequences where temporal dynamics become more complex, potentially leading to challenges such as overfitting or difficulties in temporal feature retention over extended durations.

In contrast, the ViT model demonstrated an improvement in performance with longer sequences, achieving accuracy of 92.44\% for 12 frames, 92.82\% for 18 frames, and 93.69\% for 24 frames. This ascending pattern supports the notion that ViT architectures, with their inherent self-attention mechanisms, are well-suited to managing longer sequences. The ability of ViTs to assign varying degrees of importance to different parts of the sequence likely contributes to their enhanced performance on longer input frames.

The hybrid CNN-ViT model showcased the highest and continuously improving accuracy rates across all frame lengths: 94.12\% for 12 frames, 94.56\% for 18 frames, and an impressive 95.78\% for 24 frames. Moreover, the pre-trained hybrid model showcased the same trend, with the best accuracy achieved.
This type of model synergistically combines CNN’s robust spatial feature extraction capabilities with ViT’s efficient handling of temporal relationships via self-attention. The results from this model indicate that such a hybrid approach is particularly effective in capturing the complexities of action recognition tasks in video sequences, especially as the sequence length increases.

These findings underscore the potential advantages of hybrid neural network architectures in video-based action recognition tasks, particularly for handling longer sequences with complex interactions. The superior performance of the hybrid CNN-ViT model suggests that integrating the spatial acuity of CNNs with the temporal finesse of ViTs can lead to more accurate and reliable recognition systems. Future work could explore the scalability of these models to other datasets, their computational efficiency, and their robustness against variations in video quality and scene dynamics. Additionally, further research might investigate the optimal balance of CNN and ViT components within hybrid models to maximize both performance and efficiency.

\begin{table}[htbp]
\centering
\caption{Comparison of the proposed hybrid model with the state-of-the-art models on the KTH dataset.}
\begin{tabular}{l|l|l}
\hline
\textbf{Methods} & \textbf{Venue} & \textbf{Accuracy} \\ \hline \hline
\cite{geng2016human} & ICCSAE ’16      & 92.49      \\ 
\cite{arunnehru2018human}       & RoSMa ’18      & 94.90      \\ 
\cite{abdelbaky2020human}  & ITCE ’20      & 87.52      \\ 
\cite{jaouedi2020new}      & KSUCI journal ’20      & 96.30      \\ 
\cite{liu2020construction}     & JAIHC ’20      & 91.93      \\

\cite{sahoo2020har}       & TETCI ’20     & 97.67     \\ 
\cite{lee2021video}  &  CVF ’21    & 89.40    \\ 
\cite{basha2022information}    & MTA journal ’22      & 96.53      \\ 

\cite{ye2023unified}      & CVF ’23     & 90.90      \\ \hline
Ours      & -     & \textbf{97.89}     \\ \hline
\end{tabular}
\label{table:SoAcOMPARISION}
\end{table}

To complete the comparison, Table \ref{table:SoAcOMPARISION} shows that the impressive 97.89\% accuracy achieved by the presented CNN-ViT hybrid model on the KTH dataset places it prominently among state-of-the-art models for HAR. This performance is notably superior when compared to earlier benchmarks reported in the literature such as \cite{geng2016human} with 92.49\% and \cite{arunnehru2018human} with 94.90\%. Our model utilizes an ImageNet-pre-trained MobileNet \citep{howard2017mobilenets} as the CNN backbone in the spatial component, which enhances its robust feature extraction capabilities. Combined with the dynamic attention mechanisms of ViT, it can thereby enhance both the spatial and temporal processing of video sequences.
Furthermore, our hybrid model not only surpasses other contemporary approaches like \cite{liu2020construction}  (91.93\%) and \cite{lee2021video} (89.40\%), but also shows competitive/superior performance against some of the highest accuracy in the field, such as \cite{jaouedi2020new} (96.30\%) and \cite{basha2022information} (96.53\%). Even in comparison to the high benchmark set by \cite{sahoo2020har} (97.67\%), our hybrid model demonstrates a marginal but significant improvement, underscoring the efficacy of integrating CNN with ViT. This integration not only facilitates more nuanced feature extraction across both spatial and sequential dimensions but also adapts more dynamically to the varied contexts inherent in video data, making it a potent solution for realistic action recognition scenarios.

On the whole, the integration of CNN with ViT is particularly advantageous for enhancing feature extraction capabilities and focusing on relevant segments dynamically through the attention mechanisms of ViTs. This not only helps in improving accuracy but also in making the model more adaptable to varied video contexts, a key requirement for action recognition in realistic scenarios. This comparative advantage suggests that hybrid models are paving the way for future explorations in HAR, combining the best of convolutional and ViT-based architectures for improved performance and efficiency.

\section{Challenges and Future Directions}
\label{sec:sec5}
The field of HAR faces several formidable challenges that stem from the inherent complexity of interpreting human movements within diverse and dynamic environments. One of the primary obstacles is the variability in human actions themselves, which can differ significantly in speed, scale, and execution from one individual to another \citep{pareek2021survey}. This variability necessitates the development of sophisticated models capable of generalizing across a wide range of actions without sacrificing accuracy \citep{nayak2021comprehensive}. Additionally, the presence of complex backgrounds and environments further complicates the task of HAR. Systems must be adept at isolating and recognizing human actions against a backdrop of potentially distracting or obstructive elements, which can vary from the bustling activity of a city street to the unpredictable conditions of outdoor settings \citep{wang2013action, he2016deep}.

HAR systems furthermore must navigate the fine line between inter-class similarity and intra-class variability, where actions that are similar to each other (such as running versus jogging) require nuanced differentiation, while the same action can appear markedly different when performed by different individuals or under varying circumstances \citep{gong2020learning, zhu2018compound}. The challenge of temporal segmentation adds another layer of complexity, as accurately determining the start and end of an action within a continuous video stream is crucial for effective recognition \citep{zolfaghari2018eco}. Coupled with the need for computational efficiency to process video data in real-time and the difficulties associated with obtaining large, accurately annotated datasets, these challenges underscore the multifaceted nature of HAR \citep{caba2015activitynet}. Addressing these issues is critical for advancing the field and enhancing the practical applicability of HAR systems in real-world applications, from surveillance and security to healthcare and entertainment.

The motivation behind this work has been driven by the compelling need to bridge the existing gaps between the spatial feature extraction capabilities inherent in CNNs and the dynamic temporal processing strengths found in ViTs \citep{arnab2021vivit}. Through the introduction of a novel hybrid model, an attempt has been made to leverage the synergistic potential of these technologies, thereby enhancing the accuracy and efficiency of HAR systems in capturing the complex spatial-temporal dynamics of human actions.

Looking forward, a promising future for HAR is envisioned, particularly through the development of hybrid and integrated models. It is believed that the potential of these models extends beyond immediate performance improvements, inspiring new directions for research within the field. It is anticipated that future studies will focus on optimizing these hybrid architectures, aiming to make them more scalable and adaptable to real-world applications across various domains such as surveillance, healthcare, and interactive media. Furthermore, the exploration of self-attention mechanisms and the adaptation of large-scale pre-training strategies from ViTs are seen as exciting prospects for HAR. These approaches are expected to lead to the development of more sophisticated models capable of understanding and interpreting human actions with unprecedented accuracy and nuance.

The integration of CNNs and ViTs into hybrid CNN-ViT models presents a promising avenue for overcoming the challenges faced by HAR systems. These hybrid models capitalize on the strengths of both architectures: the local feature extraction capabilities of CNNs and the global context understanding of ViTs. Future developments could focus on enhancing model adaptability to generalize across diverse actions, improving the isolation of human actions from complex backgrounds through advanced attention mechanisms, and developing nuanced differentiation techniques for closely related actions \citep{carion2020end}. Innovations in model architecture, alongside the application of transfer learning and few-shot learning techniques, could significantly reduce the variability challenge in human actions.

Moreover, addressing the temporal segmentation challenge requires the integration of specialized temporal modules and sequence-to-sequence models to accurately determine the start and end of an action within continuous video streams. Computational efficiency remains paramount for real-time processing, necessitating ongoing efforts in model optimization and the exploration of synthetic data generation to mitigate the difficulties associated with obtaining large and accurately annotated datasets. Customizable hybrid CNN-ViT models that can be tailored for specific applications, from surveillance to healthcare, will ensure that these advancements not only push the boundaries of academic research but also enhance practical applicability in real-world scenarios. Through these concerted efforts, hybrid CNN-ViT models are poised to make significant contributions to the field of HAR, offering innovative solutions to its multifaceted challenges.

This work has highlighted the importance of continued innovation and cross-disciplinary collaboration in the advancement of HAR technologies. By integrating insights from computer vision, machine learning, and domain-specific knowledge, it is hoped that HAR systems will not only become more efficient and accurate but also more responsive to the complexities and variances of human behavior in natural environments. As the field moves forward, the focus is set on pushing the boundaries of what is possible in HAR, with the aim of creating systems that enhance human-computer interaction and contribute positively to society through various applications.

%-----------------------------
\section{Conclusions}
\label{sec:sec6}

This survey provides a comprehensive overview of the current state of HAR by examining the roles and advancements of CNNs, RNNs, and ViTs. It delves into the evolution of these architectures, emphasizing their individual contributions to the field. The introduction of a hybrid model that combines the spatial processing capabilities of CNNs with the temporal understanding of ViTs represents a methodological advancement in HAR. This model aims to address the limitations of each architecture when used in isolation, proposing a unified approach that potentially enhances the accuracy and efficiency of action recognition tasks.
The paper identifies key challenges and opportunities within HAR, such as the need for models that can effectively integrate spatial and temporal information from video data. The exploration of hybrid models, as suggested, offers a pathway for future research, particularly in improving model performance on complex video datasets. The discussion encourages further investigation into optimizing these hybrid architectures and exploring their applicability across various domains. This work sets a foundation for future studies to build upon, aiming to push the boundaries of what is currently achievable in HAR and to explore new applications of these technologies in real-world scenarios.

%-----------------------------

\vspace{6pt}

\begin{comment}

\bmhead{Supplementary information}

If your article has accompanying supplementary file/s please state so here. 

Authors reporting data from electrophoretic gels and blots should supply the full unprocessed scans for key as part of their Supplementary information. This may be requested by the editorial team/s if it is missing.

Please refer to Journal-level guidance for any specific requirements.

\end{comment}

\bmhead{Acknowledgements}

K.A. and H.I.A. are thankful for the support from The Ministry of Education in Saudi Arabia and the Republic of Turkiye Ministry of National Education, respectively.

\bmhead{Author contributions} Conceptualisation, K.A. and X.C.; methodology, K.A.; software, K.A.; validation,, all authors.; investigation, all authors; resources, K.A. and H.I.A.; data curation, K.A.;
writing—original draft preparation, all authors; writing—review and editing, all authors; visualisation, K.A. and H.I.A.; supervision, X.C. All authors have read and agreed to the published version of
the manuscript.
    
\section*{Declarations}

%Some journals require declarations to be submitted in a standardised format. Please check the Instructions for Authors of the journal to which you are submitting to see if you need to complete this section. If yes, your manuscript must contain the following sections under the heading `Declarations':

%\begin{itemize}
\textbf{Competing interests} The authors declare no competing interests.

\bibliography{sn-bibliography}% common bib file

\end{document}